\documentclass[11pt]{article}

\usepackage[final]{acl}

\newcommand{\coderepourl}{https://github.com/znhoughton/llm-phrasal-compositionality}
\newcommand{\anoncoderepourl}{https://anonymous.4open.science/r/llm-phrasal-compositionality-744A/}
\newcommand{\weightsrepourl}{https://huggingface.co/znhoughton}

\newcommand{\aclmodename}{final}
\newcommand{\codefootnote}{%
  \ifnum\pdfstrcmp{\aclmodename}{review}=0
    \ifdefined\anoncoderepourl
      \footnote{All code and analyses in this paper can be found here: \url{\anoncoderepourl}.}%
    \else
      \footnote{All code and analyses will be released on acceptance; the link is
        omitted here to preserve double-blind review.}%
    \fi
  \else
    \ifdefined\coderepourl
      \footnote{All code and analyses in this paper can be found here: \url{\coderepourl}.}%
    \fi
  \fi}
\newcommand{\weightsfootnote}{%
  \ifnum\pdfstrcmp{\aclmodename}{review}=0
    \ifdefined\anonweightsrepourl
      \footnote{Model weights and checkpoints are released at \url{\anonweightsrepourl}.}%
    \else
      \footnote{A model-weights link is omitted here to preserve double-blind
        review; the weights and checkpoints are released.}%
    \fi
  \else
    \ifdefined\weightsrepourl
      \footnote{Model weights and checkpoints are released at \url{\weightsrepourl}.}%
    \fi
  \fi}

\usepackage{times}
\usepackage{latexsym}
\usepackage[T1]{fontenc}
\usepackage[utf8]{inputenc}
\usepackage{textcomp}   % euro sign and other symbols
\usepackage{microtype}
\usepackage{inconsolata}

\usepackage{amsmath}
\usepackage{amssymb}

\usepackage{longtable}
\usepackage{booktabs}
\usepackage{array}     % pandoc uses >{\raggedright\arraybackslash}p{} column types
\usepackage{calc}      % pandoc uses \real{} expressions for column widths
\usepackage{multirow}  % multi-row table cells

\usepackage{etoolbox}
\def\fnum@table{\tablename~\thetable}

\usepackage{float}

\usepackage{graphicx}
\makeatletter
\def\maxwidth{\ifdim\Gin@nat@width>\linewidth\linewidth\else\Gin@nat@width\fi}
\def\maxheight{\ifdim\Gin@nat@height>\textheight\textheight\else\Gin@nat@height\fi}
\def\fps@figure{htbp}
\makeatother

\IfFileExists{upquote.sty}{\usepackage{upquote}}{}  % straight quotes in verbatim
\usepackage{fancyvrb}
\date{}

\makeatletter
\newsavebox\pandoc@box
\newcommand*\pandocbounded[1]{%
  \sbox\pandoc@box{#1}%
  \Gscale@div\@tempa{\linewidth}{\wd\pandoc@box}%
  \ifdim\@tempa\p@<\p@\scalebox{\@tempa}{\usebox\pandoc@box}\else#1\fi%
}
\makeatother

\title{The Holistic Storage of Verb+\emph{Up} Phrases in Text-based and
Audio-based Language Models}

\author{
  Zachary N. Houghton \\
      University of Oregon \\ Vail Systems, Inc \\
        \texttt{znh@uoregon.edu}
     \And
  Yu Zhou \\
      Vail Systems, Inc \\
       \And
  Dan Pluth \\
      Vail Systems, Inc \\
       \And
  Jordan Hosier \\
      Vail Systems, Inc \\
       \And
  Vijay K. Gurbani \\
      Vail Systems, Inc \\
      }

\usepackage{booktabs}
\usepackage{longtable}
\usepackage{array}
\usepackage{multirow}
\usepackage{wrapfig}
\usepackage{float}
\usepackage{colortbl}
\usepackage{pdflscape}
\usepackage{tabu}
\usepackage{threeparttable}
\usepackage{threeparttablex}
\usepackage[normalem]{ulem}
\usepackage{makecell}
\usepackage{xcolor}
\makeatletter
\@ifpackageloaded{caption}{}{\usepackage{caption}}
\AtBeginDocument{%
\ifdefined\contentsname
  \renewcommand*\contentsname{Table of contents}
\else
  \newcommand\contentsname{Table of contents}
\fi
\ifdefined\listfigurename
  \renewcommand*\listfigurename{List of Figures}
\else
  \newcommand\listfigurename{List of Figures}
\fi
\ifdefined\listtablename
  \renewcommand*\listtablename{List of Tables}
\else
  \newcommand\listtablename{List of Tables}
\fi
\ifdefined\figurename
  \renewcommand*\figurename{Figure}
\else
  \newcommand\figurename{Figure}
\fi
\ifdefined\tablename
  \renewcommand*\tablename{Table}
\else
  \newcommand\tablename{Table}
\fi
}
\@ifpackageloaded{float}{}{\usepackage{float}}
\floatstyle{ruled}
\@ifundefined{c@chapter}{\newfloat{codelisting}{h}{lop}}{\newfloat{codelisting}{h}{lop}[chapter]}
\floatname{codelisting}{Listing}

\makeatother
\makeatletter
\@ifpackageloaded{caption}{}{\usepackage{caption}}
\@ifpackageloaded{subcaption}{}{\usepackage{subcaption}}
\makeatother

\begin{document}
\maketitle

\begin{abstract}
One of the most central aspects of language processing is the ability to
trade off between stored representations and abstract knowledge: one
must retrieve stored representations, but also generate novel ones by
applying productive rules. While recent work has examined abstract
knowledge in language models, holistic storage has received far less
attention. We probe internal representations in both text-based LLMs and
an ASR model, testing whether V+\emph{up} phrasal verbs develop distinct
representations as a function of frequency and predictability. All
models show evidence of holistic storage driven by frequency and
predictability, further supporting usage-based theories of language.
\end{abstract}

% Float tuning carried over from the posted arXiv build, so this lays out the same way.
% Without these, figures and tables migrate away from their declaration points and the
% appendix in particular repaginates. Kept here rather than in a separate header file:
% .gitignore tracks only writeup.qmd under paper/, so an extra file would not survive a
% clone. These are all counter/length/macro assignments, so they apply from here on.
\renewcommand{\topfraction}{0.9}
\renewcommand{\textfraction}{0.07}
\renewcommand{\floatpagefraction}{0.7}
\setcounter{topnumber}{1}
\setcounter{bottomnumber}{1}
\setcounter{totalnumber}{2}
\setlength{\floatsep}{36pt plus 6pt minus 6pt}
% Main-text figures: top/bottom/page, never "here". Switched to htbp after \appendix.
\floatplacement{figure}{tbp}

\section{Introduction}\label{introduction}

A central debate in linguistics concerns how humans trade off between
computation and storage
\citep{stembergerAreInflectedForms2004, stembergerFrequencyLexicalStorage1986, kapatsinski2009frequency, houghton2023doespredictabilitydrive, houghton2024frequencydependentpreferenceextremity, houghton2025boltsnutslanguage, houghton2025multiwordrepresentationsminds, houghton2026exemplarsdisguisepure, morgan2016abstractknowledgedirect, morgan2024productiveknowledgeitemspecific, morganModelingIdiosyncraticPreferences2015}.
Computation refers to applying abstract knowledge to generate new
representations; for example, deriving \emph{wugs} from \emph{wug} via a
productive pluralization rule. Storage refers to retrieving a whole
representation from memory rather than computing it, such as accessing a
holistic representation of a common phrase (e.g., \emph{I don't know})
despite being able to, in principle, generate the phrase
compositionally. Both mechanisms are clearly at work in language
learning and processing, but the factors that drive what forms are
produced via computation and what forms are stored and retrieved
holistically remains poorly understood.

\subsection{Computation vs Storage in
Humans}\label{computation-vs-storage-in-humans}

There is an abundance of evidence for abstract knowledge in language use
by humans. Children generalize morphological rules productively to novel
words they have never encountered
\citep{berkoChildsLearningEnglish1958}, and priming studies show that
sentences sharing syntactic structure or semantically related words
facilitate processing of sentences sharing that structure, implying that
something abstract is shared across their representations
\citep{bock1986syntacticpersistencelanguage, meyer1971facilitationrecognizingpairs}.
Similarly, when ordering novel or low-frequency binomials, humans show
sensitivity to abstract preferences (e.g., placing the shorter word
first), rather than simply favoring the more frequent ordering
\citep{morgan2016abstractknowledgedirect}.

Storage has had a more controversial history. Early accounts held that
only irregular forms (e.g., \emph{went}) were stored holistically and
that any compositionally derivable form was derived via computation
\citep{pinkerFutureTense2002}. Evidence against this view has since
accumulated
\citep{kapatsinski2009frequency, bybeeEffectUsageDegrees1999, morgan2016abstractknowledgedirect, houghton2025multiwordrepresentationsminds, stembergerAreInflectedForms2004}.
\citet{stembergerAreInflectedForms2004} showed that inflection errors
are less common for high-frequency words, suggesting holistic storage
even for regularly derived forms. \citet{bybeeEffectUsageDegrees1999}
demonstrated that \emph{don't} is phonetically more reduced in
high-frequency phrases like \emph{I don't know} than in lower-frequency
phrases like \emph{I don't go}; if \emph{don't} always had the same
representation, such context-specific reduction would be difficult to
explain.

Processing studies have provided converging evidence.
\citet{morgan2016abstractknowledgedirect} found that while human
ordering preferences for low-frequency binomials are driven by abstract
preferences, preferences for high-frequency binomials are driven by
item-specific preferences, suggesting a frequency-dependent shift from
computation to storage. Most directly relevant to the present study,
\citet{kapatsinski2009frequency} used a recognition paradigm in which
participants responded upon hearing \emph{up} within a word or a
V+\emph{up} phrase, and found that reaction times were U-shaped across
frequency: reaction times were faster for medium-frequency V+\emph{up}
phrases than low-frequency ones, but slower again for high-frequency
phrases. Further, \citet{houghton2025multiwordrepresentationsminds}
found that this pattern holds for high-predictability phrases
(V+\emph{up} phrases where \emph{up} is likely to appear given the
verb), suggesting that high-frequency and high-predictability
V+\emph{up} phrases are represented holistically.

\subsection{Computation vs Storage in
LMs}\label{computation-vs-storage-in-language-models}

Whether language models exhibit analogous computation-storage tradeoffs
to humans has become an active area of inquiry. On the computation side,
results are mixed: some studies find that models learn abstract
generalizations not present in training
\citep{misraLanguageModelsLearn2024, yaoBothDirectIndirect2025}, while
others find that models fail to use abstract knowledge where humans do,
such as morphological generalization to novel words
\citep{haleyThisBERTNow2020}. On the storage side, there is no doubt
that models rely heavily on memorization
\citep{mccoy2023howmuchlanguage}. Indeed, item-specific frequency
effects have been documented even in tasks where humans show abstract
preferences \citep{houghton2025roleabstractrepresentations}.

Despite these findings, it remains unclear as to whether LLMs develop
holistic phrasal representations in a manner similar to humans. If they
do, this would suggest that holistic storage is a natural consequence of
learning from distributional patterns in the language, requiring no
explicit storage mechanism, and lending support to usage-based accounts
of how computation and storage interact. Audio-based models are
especially well-suited to address this question because a good deal of
the evidence for holistic storage comes from listening paradigms. It is
thus important to understand whether Automatic Speech Recognition (ASR)
models, not just LLMs, show evidence of holistic storage; yet holistic
storage has never been examined in any ASR model \citep[though see][ for
related mechanistic interpretability work on Whisper, an ASR
model]{pluth2026mechanisticinterpretabilityasrmodels}, and has rarely
been examined in any modality or in models trained on human-comparable
amounts of data.

\subsection{Present Study}\label{present-study}

The present study addresses this gap. We probe internal representations
of text-based language models that were trained on an amount of data
comparable to humans (BabyLMs),\footnote{The model was trained on 150
  million tokens. The average college-aged human experiences
  approximately 350 million words \citep{levy2012processing} so the
  model is trained on a little less than half the tokens that an average
  college-aged human has experienced.} a large language model (OLMo-3
7B), and an ASR model (Whisper-small). These models were chosen to help
illuminate effects of training size, number of parameters, and modality
(speech vs text). In order to probe their representations, we trained
logistic classifiers to detect the embedding of \emph{up} as a
preposition (outside of V+\emph{up} contexts), then tested the
classifier on V+\emph{up} phrases of varying frequency and
predictability. If these models develop holistic representations for
V+\emph{up} phrases, the representation of \emph{up} in high-frequency
and high-predictability V+\emph{up} phrases should diverge from that of
the preposition \emph{up} more than in lower-frequency and
lower-predictability phrases, resulting in lower logit scores by the
classifier. Our specific contributions are:

\begin{itemize}
\item
  We train and release three open-access autoregressive models trained
  on the BabyLM v3 corpus \citep{charpentier2025babylm}, checkpointed
  every 20M tokens, to facilitate future research on human-scale
  language learning.\weightsfootnote
\item
  We show that holistic phrasal storage emerges in both text-based LLMs
  and an ASR model, establishing that frequency- and
  predictability-driven holistic representations arise even from models
  trained on an amount of data comparable to humans, and even across
  modalities.
\item
  We show that frequency effects on phrasal storage are robust across
  model sizes, but predictability effects strengthen with scale,
  suggesting that sensitivity to co-occurrence statistics beyond raw
  frequency requires greater representational capacity.
\end{itemize}

\section{Model Training}\label{model-training}

In order to examine holistic storage in models that have seen
human-comparable amount of data, we trained three language models on the
BabyLM v3 corpus \citep{charpentier2025babylm}, a 150M-token dataset
designed to be more reflective of the scale and quality of the language
that humans receive.\footnote{Though it is worth noting, as has been
  pointed out before
  \citep[e.g.,][]{houghton2025multiwordrepresentationsminds}, that it
  may be misleading to compare the tokens that LLMs receive to the
  ``tokens'' that humans receive, since humans encounter language in a
  context-rich environment while LLMs see only the raw text.} All three
models follow the OPT decoder-only transformer architecture
\citep{zhang2022opt}, with 125M, 350M, and 1.3B parameters,
respectively.\codefootnote

Prior to training, we fit a byte-pair encoding (BPE) tokenizer directly
on the BabyLM training corpus, yielding a vocabulary of 8,192 subword
types. This tokenizer was shared across all three model sizes, ensuring
that cross-model comparisons are not confounded by differences in
tokenization. A full description of the model training is included in
Appendix~\ref{babylm-model-training}; training loss curves, final
gradient norms, and validation perplexity confirming that all three
models converged are reported in
Appendix~\ref{babylm-model-convergence}.

\section{Experiment 1: UP
Independently}\label{experiment-1-up-independently}

Experiment 1 tests whether LLMs develop holistic phrasal representations
analogous to those proposed by usage-based theories
\citep[e.g.,][]{houghton2025multiwordrepresentationsminds}, using
classifiers trained on prepositional \emph{up} representations and
applied to V+\emph{up} phrases varying in frequency and predictability.
We investigate this across OLMo-3 7B and the three BabyLMs (OPT-125M,
350M, and 1.3B).

\subsection{Methods}\label{methods}

For each model, we extracted the representation of the token \emph{up}
from each hidden layer of the models for each sentence. Using these
representations, we trained a separate logistic regression classifier
for each layer to distinguish prepositional \emph{up} (which did not
occur in V+\emph{up} phrases in the training of the classifier) from
other tokens in the same sentence (these other tokens did not contain
the segment \emph{up} in any capacity). The trained classifier was then
tested on a held-out test set comprising V+\emph{up} phrases, and for
each sentence the classifier returned a logit score reflecting the
predicted probability that the representation of \emph{up} in the
V+\emph{up} phrase resembled the standalone, prepositional class.

Frequency for each V+\emph{up} type is operationalized as its raw corpus
count, log-transformed. Letting
\(c_{vup} = \text{count}(V\text{+\emph{up}})\):

\begin{equation}\phantomsection\label{eq-log-freq}{\text{log-frequency} = \log\bigl(c_{vup}\bigr)}\end{equation}

Predictability is operationalized as the log-odds ratio of V+\emph{up}
occurrences to V not followed by up in the corpus.\footnote{Note that
  this is mathematically equivalent to taking the logit of the
  conditional probability of \emph{up} given the verb.} Letting
\(c_V = \text{count}(V)\):

\begin{equation}\phantomsection\label{eq-logit-predic}{\text{log-predictability} = \log\!\left(\frac{c_{vup}}{c_V - c_{vup}}\right)}\end{equation}

Counts were derived from the training corpus of each model: the BabyLM
V3 dataset \citep{charpentier2025babylm} for the BabyLM models, and
Dolma v1.7 (queried via the infini-gram API; \citealt{liu2024infini})
for OLMo-3 7B. Although OLMo-3 7B was trained on Dolma 3, Dolma v1.7 is
the most recent snapshot indexed by infini-gram and draws from the same
underlying sources, making it a reasonable approximation of OLMo-3 7B's
training distribution.\footnote{The relative frequencies of English
  phrasal verbs are unlikely to differ substantially between Dolma
  versions, as both are large-scale web-text corpora of similar
  composition.}

\subsubsection{Classifier Training}\label{classifier-training}

The classifier was trained to distinguish the language models'
representations of the preposition \emph{up} from representations of
other tokens in the same sentence. Positive training examples consisted
of 1,000 occurrences of \emph{up} which occurred in sentences strictly
as a preposition,\footnote{We used a morphological parser to filter out
  sentences in which \emph{up} was not tagged as a preposition. The
  filter is deliberately conservative: it requires the unambiguous
  dependency label \texttt{prep}.} drawn from sentences in the C4 corpus
\citep{raffel2020exploring}. Because \emph{up} is the only word among
these positive examples, this set is simply 1,000 different sentences
containing that same word, not 1,000 different words. Negative examples
were 1,000 tokens randomly selected from the same sentences, restricted
to tokens whose decoded string consists entirely of alphabetic
characters (no numbers, punctuation, or special characters), and
excluding the preposition \emph{up} itself and any token containing
\emph{up} as a substring. The validation set was drawn from the same
pool of sentences (a non-overlapping subset), with token positions
resolved per model's tokenizer; exact counts vary slightly by tokenizer
(Appendix~\ref{experiment-1-up-independently-1}). The classifiers for
all models were trained on the same underlying sentences.\footnote{Across
  all layers, models, and experiments, every classifier achieved above
  94\% accuracy on its held-out validation set. A full per-layer
  breakdown is reported in Appendix~\ref{app-val-accuracy}.}

The test set comprised V+\emph{up} phrases (e.g., \emph{pick up}) with
at least 20 occurrences in the corpus; up to 20 sentences were sampled
per type. Because the BabyLM corpus is smaller than Dolma, fewer
V+\emph{up} types attain a valid (non-zero) predictability estimate;
only types with a valid estimate are included in the analyses. Full
item-level statistics are reported in Appendix~\ref{test-set-items}
(Table~\ref{tbl-test-items}), but broadly speaking there were 4,081
unique V+\emph{up} types for OLMo-3 and 1,039 for the BabyLM models.
Classifier training and validation split sizes are shown in
Appendix~\ref{experiment-1-up-independently-1}
(Table~\ref{tbl-exp1-splits}).

\subsubsection{Analyses}\label{analyses}

In order to examine the effects of frequency and predictability on the
representations of \emph{up} in V+\emph{up} phrases, we implemented a
Bayesian mixed-effects regression model using the \emph{brms} package
\citep{burknerBrmsPackageBayesian2017}. For each statistical analysis,
the outcome variable was the logit score returned by the classifier on
each test item. We included fixed-effects for frequency and
predictability, both of which were centered and scaled (denoted
\(c\_\text{log\_freq}\) and \(c\_\text{log\_predic}\) below). We also
included random intercepts for phrasal verb type. Weak, uninformative
priors were included on each fixed-effect. The model syntax is included
below:

\begin{equation}\phantomsection\label{eq-linear-joint}{
\begin{aligned}
\text{logit} &\sim c\_\text{log\_freq} * c\_\text{log\_predic} \\
             &\quad + (1 \mid \text{verb\_up})
\end{aligned}
}\end{equation}

Frequency and predictability are correlated to some degree, but because
both are included in the same model, each coefficient reflects effects
above and beyond the other, and variance inflation factors remain low
across all models (Appendix~\ref{test-set-items}).

Separate statistical models were fit for each language model at the
final hidden layer. For each coefficient we report the posterior mean,
standard error, 95\% credible interval (CI), and the proportion of
posterior samples greater than zero (\% \textgreater{} 0). We consider
an effect to be meaningful if the 95\% CI does not contain
zero.\footnote{Though as \citet{houghton2024task} point out, unlike
  frequentist statistics, Bayesian analyses don't force us to commit to
  a binary of significance vs.~non-significance, and the percentage of
  samples greater than zero can be interpreted in a continuous manner.}

In order to examine the difference in representations across hidden
layers, we additionally fit a generalized additive model
\citep[GAM,][]{wood2017generalized}. As with the Bayesian analyses, the
dependent variable was the classifier logit. Specifically, two models
were fit separately, one for frequency and one for predictability, each
with a tensor product smooth over the predictor and hidden layer index
(with a random intercept for verb):

\begin{equation}\phantomsection\label{eq-gam-layers}{
\begin{aligned}
\text{logit} &\sim \mathrm{te}(\text{log\_freq},\, \text{layer}) \\
             &\quad + s(\text{verb\_up},\, \text{bs}{=}\texttt{`re'}) \\[4pt]
\text{logit} &\sim \mathrm{te}(\text{log\_predic},\, \text{layer}) \\
             &\quad + s(\text{verb\_up},\, \text{bs}{=}\texttt{`re'})
\end{aligned}
}\end{equation}

\subsection{Results}\label{results}

\begin{figure}

\centering{

\pandocbounded{\includegraphics[keepaspectratio]{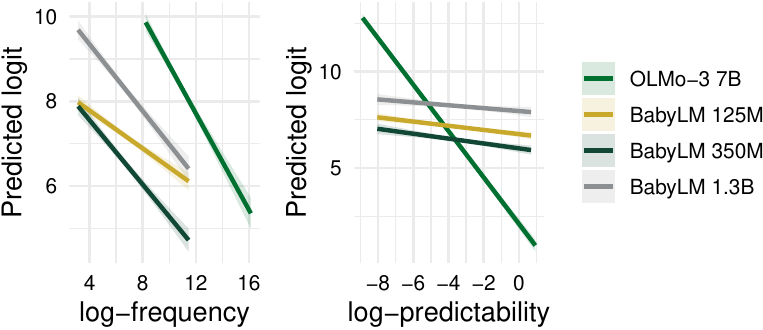}}

}

\caption{\label{fig-exp1-joint-surfaces}Final-layer brms predicted logit
by frequency (left) and predictability (right) for all models (UP
independently). Shading indicates 95\% CIs.}

\end{figure}

\begin{figure}

\centering{

\pandocbounded{\includegraphics[keepaspectratio]{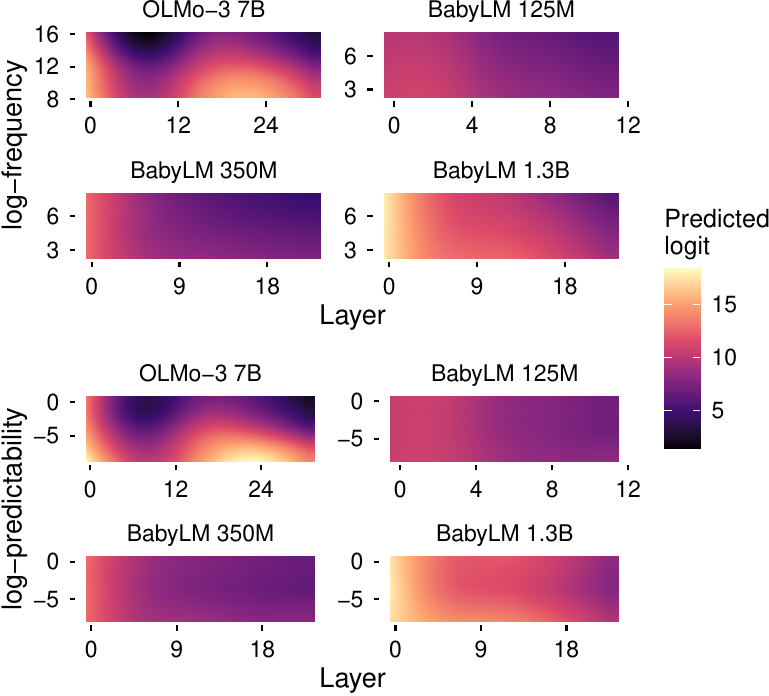}}

}

\caption{\label{fig-exp1-gam-layers}GAM-predicted logit by layer and
predictor for all models (UP independently). Top: frequency; bottom:
predictability.}

\end{figure}

Full numerical results are reported in
Appendix~\ref{experiment-1-up-independently-1}
(Table~\ref{tbl-exp1-joint-final} for the Bayesian model and
Table~\ref{tbl-exp1-gam-smooth} for the GAM); they are visualized in
Figure~\ref{fig-exp1-joint-surfaces} and
Figure~\ref{fig-exp1-gam-layers}. Figure~\ref{fig-exp1-gam-diffs}
(Appendix~\ref{experiment-1-up-independently-1}) shows the
layer-by-layer difference between high- and low-predictor items more
directly.

For all four LLMs, there was a negative effect of both frequency and
predictability on the classifier's logit score. In other words, both
frequency and predictability result in the representation of \emph{up}
in V+\emph{up} phrases being less similar to the prepositional
representation of \emph{up}. We also observe a negative
frequency-by-predictability interaction for all three BabyLM models but
not for OLMo-3 7B (Table~\ref{tbl-exp1-joint-final}).

Additionally, the by-layer analysis demonstrates that predictability's
divergence begins to appear early on in the larger models (specifically,
OLMo-3 7B and BabyLM 1.3B), while taking longer to emerge in the smaller
models.

\subsection{Discussion}\label{discussion}

Experiment 1 found that the logits from a classifier trained on the
preposition \emph{up} representations decreased for high-frequency and
high-predictability V+\emph{up} phrases. Additionally, the effect of
predictability appears to be weaker in the smaller models. These results
parallel those of \citet{kapatsinski2009frequency} and
\citet{houghton2025multiwordrepresentationsminds}, who examined this in
humans. In Experiment 2, adopt a more similar approach to theirs by also
examining a probe trained to identify the \emph{up} segment, regardless
of semantics.

\section{Experiment 2: UP as Subword}\label{experiment-2-up-as-subword}

One possible explanation for the results found in Experiment 1 is that
the lower logit score for frequent/predictable V+\emph{up} phrases could
reflect semantic bleaching or polysemy rather than holistic storage. To
address this concern, we replicate the design of Experiment 1 with one
key change: rather than training the classifier on the preposition
\emph{up} exclusively, we additionally train it on instances of
\emph{up} embedded within a larger word (e.g., \emph{cup}), requiring
the model to identify the sequence \emph{up} regardless of whether it
functions as a preposition. This tests whether the frequency and
predictability effects observed in Experiment 1 generalize to the
\emph{up}-segment more broadly. This design is also arguably more
faithful to \citet{kapatsinski2009frequency} and
\citet{houghton2025multiwordrepresentationsminds}, where participants
were tasked with recognizing the segment \emph{up} in general.

\subsection{Methods}\label{methods-1}

The procedure was identical to Experiment 1 (same models; classifiers
trained and evaluated at each hidden layer; same test set, outcome
variable, and statistical approach) with one difference: the classifier
was trained on a broader set of positive examples that included
instances of \emph{up} embedded within a larger word (e.g., \emph{cup},
\emph{puppy}).

\subsubsection{Classifier Training}\label{classifier-training-1}

The training set for this experiment combined two types of positive
examples: 1,000 occurrences of the preposition \emph{up} (identical to
Experiment 1) and 1,000 occurrences of \emph{up} embedded within a
larger word (e.g., \emph{cup}, \emph{puppy}, etc). Critically, the
subword positives were restricted to \textbf{unique word types}: each
up-containing word contributed exactly one instance, so the classifier
could not learn to recognize a high-frequency form like \emph{cup} from
repeated exposure. Negative examples (1,000 drawn from each sentence
pool) were tokens consisting entirely of alphabetic characters that did
not contain \emph{up} as a substring. The validation set was constructed
by the same procedure (targeting 1,000 positive, 1,000 negative per
sentence pool; exact counts vary by tokenizer). As in Experiment 1, each
model used the same underlying sentences with their own tokenizers. The
test set was identical to Experiment 1. Classifier training/validation
split sizes are shown in Appendix~\ref{experiment-2-up-as-subword-1}
(Table~\ref{tbl-exp2-splits}).

\subsubsection{Analyses}\label{analyses-1}

Analyses were identical to those of Experiment 1.

\subsection{Results}\label{results-1}

\begin{figure}

\centering{

\pandocbounded{\includegraphics[keepaspectratio]{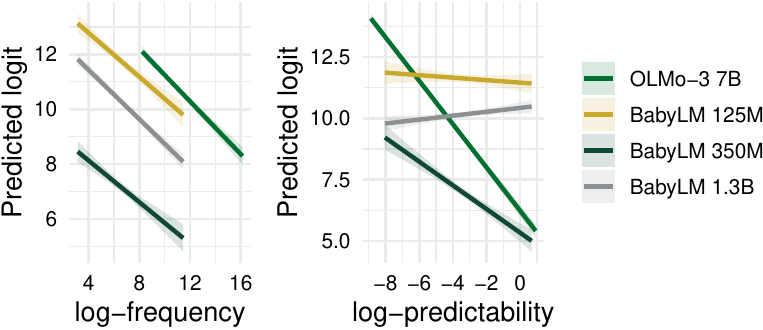}}

}

\caption{\label{fig-exp2-joint-surfaces}Final-layer brms predicted logit
by frequency (left) and predictability (right) for all models (UP as
subword). Shading indicates 95\% CI.}

\end{figure}

\begin{figure}

\centering{

\pandocbounded{\includegraphics[keepaspectratio]{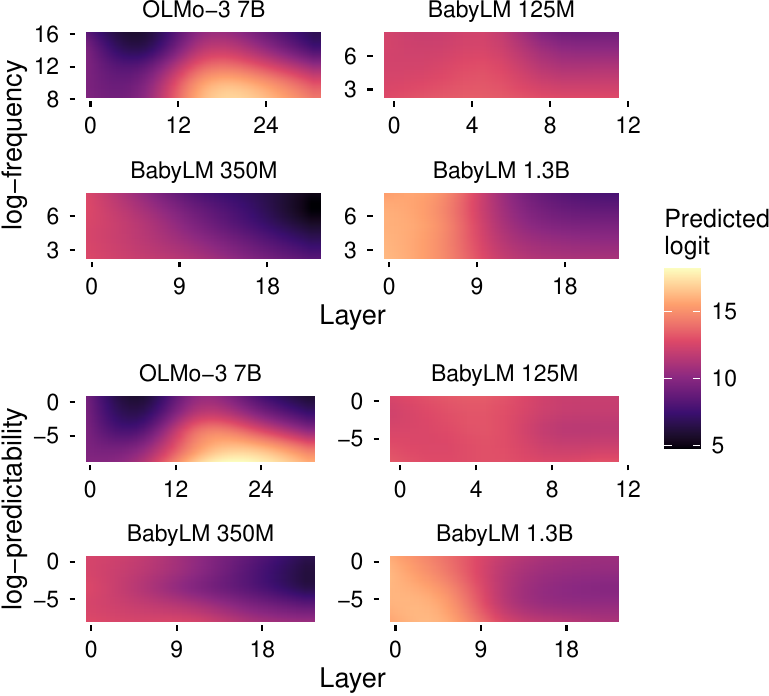}}

}

\caption{\label{fig-exp2-gam-layers}GAM-predicted logit as a function of
transformer layer (x-axis) and predictor value (y-axis) for each model
(UP as subword). Top row: log-frequency; bottom row: log-predictability.
Color encodes the predicted logit score (random effects excluded).}

\end{figure}

Full numerical results are reported in
Appendix~\ref{experiment-2-up-as-subword-1}
(Table~\ref{tbl-exp2-joint-final} for the Bayesian model and
Table~\ref{tbl-exp2-gam-smooth} for the GAMs); they are visualized in
Figure~\ref{fig-exp2-joint-surfaces} and
Figure~\ref{fig-exp2-gam-layers}. Figure~\ref{fig-exp2-gam-diffs}
(Appendix~\ref{experiment-2-up-as-subword-1}) shows the layer-by-layer
difference more directly. Frequency effects replicate Experiment 1
across all four models. Predictability effects are more variable: BabyLM
125M shows no effect, 350M shows a negative effect, and the 1.3B model
shows a positive effect. OLMo-3 7B, however, shows a consistent negative
effect, suggesting that sensitivity to predictability may increase with
scale of the model (though it's unclear whether this is due to scale of
the model or scale of the data). Similar to Experiment 1, there is also
a negative interaction effect between frequency and predictability,
however this time only credible for the two smaller BabyLMs (125M and
350M). The by-layer patterns similarly show a progressively stronger
effect of frequency for later layers relative to earlier layers, while
the effect of predictability follows a similar pattern for OLMo, but not
for the BabyLMs. For the BabyLMs, the effect of predictability either
stays relatively constant across layers or, in the case of the 1.3B
model, becomes negative around layer 5 and then slowly grows positive
again.

\subsection{Discussion}\label{discussion-1}

The results of Experiment 2 mostly replicate those of Experiment 1.
Frequency shows a consistent effect such that the representation of
\emph{up} in high-frequency V+\emph{up} phrases diverges from
representations of \emph{up}, even for a probe trained to identify
\emph{up} as a segment, regardless of the semantics. Predictability
behaves less consistently than frequency. The three BabyLMs share a
training dataset and differ only in size, yet their predictability
effects are non-monotonic: the 350M shows the strongest negative effect
of the three and the 1.3B model in contrast shows a positive effect. The
OLMo model, on the other hand, shows a large negative effect. Since OLMo
differs from the BabyLMs in both number of parameters and data
diversity, it's difficult to attribute that effect to either one alone.
Further, the results cannot be due to poor classifier training, since
the probe achieves over 97\% accuracy across models and layers at
identifying the \emph{up} segment within words
(Appendix~\ref{app-val-accuracy}). Finally, the results mitigate
concerns that Experiment 1's findings reflect polysemy: Experiment 2's
positive class is defined purely orthographically, regardless of
semantics, yet the effects of frequency and predictability largely
replicate.

\section{Experiment 3: Whisper (ASR
Model)}\label{experiment-3-whisper-asr-model}

In Experiment 3, we replicate Experiment 1 with Whisper-small, an ASR
model trained on spoken audio. Whisper differs from the models in
Experiments 1 and 2 in both its training modality and its
encoder-decoder architecture. Examining Whisper allows us to ask whether
the frequency and predictability effects on phrasal representations
generalize beyond text-based models to representations learned from
speech.

\subsection{Methods}\label{methods-2}

Experiment 3 has two key differences: (1) the model was Whisper-small;
and (2) the stimuli were spoken audio segments. For each segment, we ran
the audio through Whisper and extracted the hidden-state representation
of \emph{up} at each layer of the encoder and decoder separately. For
the encoder, this is the mean of the hidden states over the frames
spanning the aligned \emph{up} interval, and is therefore derived from
the acoustic signal alone; for the decoder, it is the hidden state at
the \emph{up} token position, with the ground-truth transcript supplied
as decoder input. A logistic regression classifier was trained
independently for each layer of the encoder and decoder to differentiate
Whisper's embeddings of \emph{up} as a standalone preposition from
non-\emph{up} tokens, then applied to spoken V+\emph{up} phrases; an
extension of this Experiment with the probe trained on words containing
\emph{up} as well is presented separately as a robustness check
(Appendix~\ref{app-whisper-subword}).

\subsubsection{Classifier Training}\label{classifier-training-2}

The Whisper experiment used spoken language data drawn from a subset of
the GigaSpeech corpus \citep{GigaSpeech2021} (corpus construction
statistics in Appendix~\ref{experiment-3-whisper}). Word-level
timestamps within each segment were obtained using WhisperX forced
alignment. In order to classify \emph{up} accurately, we used the
ground-truth transcripts of the audio segments rather than the Whisper
transcribed ones (to avoid any potential confound). We classified
\emph{up} by running spaCy part-of-speech tagging on the ground-truth
transcript: if the token immediately preceding \emph{up} was tagged as a
VERB, the instance was labelled \emph{V+up} (e.g., \emph{pick up},
\emph{clean up}). This classification procedure mirrors the one used to
identify the preposition \emph{up} in Experiment 1.

The \textbf{training and validation sets} used 1,000 preposition
\emph{up} instances each as positive examples, paired with 1,000
randomly selected non-\emph{up} words from the same segment as negative
examples.

The \textbf{test set} comprised V+\emph{up} types with at least 5
occurrences in the audio dataset (lower than Experiments 1 and 2,
reflecting the relative sparsity of the audio domain), with up to 20
instances per type. Corpus frequency and predictability were estimated
using Dolma v1.7 (via infini-gram). After filtering to items with a
valid predictability estimate, 1,460 unique V+\emph{up} types were
retained. Split sizes are described in
Appendix~\ref{experiment-3-whisper} (Table~\ref{tbl-exp3-splits}) and
item-level statistics in Appendix~\ref{test-set-items}
(Table~\ref{tbl-test-items}).

\subsubsection{Analyses}\label{analyses-2}

Analyses followed the same procedure as Experiments 1 and 2. Because
audio segments, unlike text tokens, can vary in acoustic duration, we
additionally included the duration of the \emph{up} segment
(\(c\_\text{duration}\), centered and scaled) as a covariate in the
model, to control for the phonetic reduction of \emph{up} in frequent
V+\emph{up} phrases:

\begin{equation}\phantomsection\label{eq-linear-joint-duration}{
\begin{aligned}
\text{logit} &\sim c\_\text{log\_freq} * c\_\text{log\_predic} \\
             &\quad + c\_\text{duration} + (1 \mid \text{verb\_up})
\end{aligned}
}\end{equation}

The brms models were fit separately for Whisper's encoder and decoder at
the final hidden layer. The by-layer GAMs likewise included duration as
an additional covariate:

\begin{equation}\phantomsection\label{eq-gam-layers-duration}{
\begin{aligned}
\text{logit} &\sim \mathrm{te}(\text{log\_freq},\, \text{layer}) + c\_\text{duration} \\
             &\quad + s(\text{verb\_up},\, \text{bs}{=}\texttt{'re'}) \\[4pt]
\text{logit} &\sim \mathrm{te}(\text{log\_predic},\, \text{layer}) + c\_\text{duration} \\
             &\quad + s(\text{verb\_up},\, \text{bs}{=}\texttt{'re'})
\end{aligned}
}\end{equation}

\subsection{Results}\label{results-2}

Full numerical results are reported in
Appendix~\ref{experiment-3-whisper} (Table~\ref{tbl-exp3-joint-final}
for the Bayesian model and Table~\ref{tbl-exp3-gam-smooth} for the GAM
model); they are visualized in Figure~\ref{fig-exp3-joint-surfaces} and
Figure~\ref{fig-exp3-gam-layers} below.\footnote{As a further robustness
  check, we replicated this design with a broader positive training
  class that also included \emph{up} embedded within other words,
  following Experiment 2's subword criteria adapted for audio. The
  decoder's results parallel those reported here; the encoder's
  predictability effect likewise replicates, and its frequency effect
  replicates in direction but emerges as credible
  (Appendix~\ref{app-whisper-subword}).} Figure~\ref{fig-exp3-gam-diffs}
(Appendix~\ref{experiment-3-whisper}) shows the layer-by-layer
difference between high- and low-predictor items more directly.

The results for Whisper diverge between the two components. The decoder
shows a negative effect of both frequency and predictability,
controlling for the duration of the \emph{up} segment, consistent with
the pattern found throughout the paper; both effects remain credible
even with duration included in the model, providing evidence against
phonetic reduction as a competing explanation. The encoder shows a
different pattern: predictability has a negative effect whose 95\%
credible interval marginally includes zero (94.5\% of the posterior mass
falls below zero), while frequency shows a marginally positive effect
(91.0\% of the posterior mass above zero). Duration itself has a
credible negative effect in the encoder and a credible positive effect
in the decoder.

\begin{figure}

\centering{

\pandocbounded{\includegraphics[keepaspectratio]{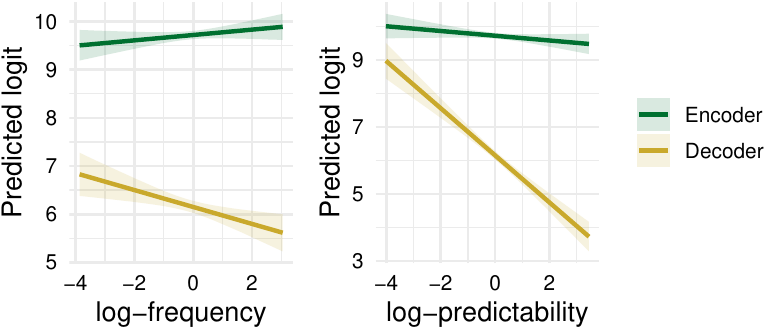}}

}

\caption{\label{fig-exp3-joint-surfaces}Final-layer brms predicted logit
by frequency (left) and predictability (right) for Whisper encoder and
decoder. Shading indicates 95\% CI.}

\end{figure}%

\begin{figure}

\centering{

\pandocbounded{\includegraphics[keepaspectratio]{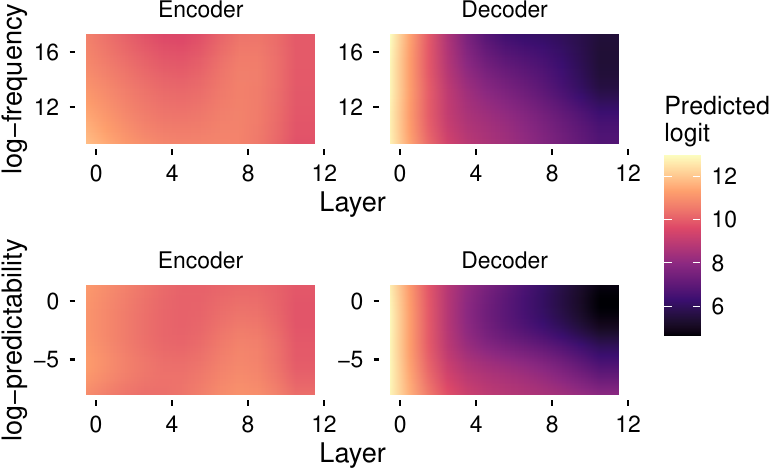}}

}

\caption{\label{fig-exp3-gam-layers}GAM-predicted logit by layer and
predictor for Whisper encoder and decoder. Top: frequency; bottom:
predictability. Color = logit; darker = lower logit.}

\end{figure}%

The by-layer analysis shows the encoder's effects are variable across
layers: predictability stays negative throughout, most pronounced in the
middle layers, while frequency stays negative through the early and
middle layers before rising sharply in the final third of the network
and crossing to a small positive difference by the final layers,
consistent with the final-layer joint model's near-zero frequency effect
reported above. In the decoder, frequency drops sharply from near zero
to strongly negative by the middle layers and remains strongly negative
through the final layer, while predictability decreases steadily and
monotonically across the entire network, reaching its most negative
value at the final layer (Table~\ref{tbl-exp3-gam-smooth}).

\subsection{Discussion}\label{discussion-2}

The results of Whisper demonstrate an interesting similarity between ASR
models and LLMs: both the encoder and the decoder of Whisper show
similar patterns to the LLMs for predictability, and the decoder shows
the same pattern for frequency as well. In the encoder, however,
frequency's relationship is positive rather than negative. The by-layer
trajectory offers a possible explanation, distinct from phonetic
reduction (which the duration covariate already controls for): the
encoder's frequency effect is negative in early layers and crosses over
to positive by the final layers (Figure~\ref{fig-exp3-gam-diffs}). One
interpretation is that the encoder's early layers retain acoustically
fine-grained detail, on which frequent instances of \emph{up} may simply
be more acoustically variable than the relatively consistent
prepositional form, while later layers progressively abstract over this
acoustic variability. Under this view, the crossover reflects the
encoder abstracting away surface acoustic differences.

\section{Conclusion}\label{discussion-and-conclusion}

The present study demonstrates that both text-based LLMs and ASR models'
representations of \emph{up} in frequent/predictable V+\emph{up} are
less similar to the representation of standalone \emph{up}, suggesting
that they are represented holistically.\footnote{Notably, all the
  classifiers showed a high accuracy on a held-out validation set,
  ruling out poor classifier training as a possible explanation for
  these results.} This result holds even in models trained on an amount
of data comparable to humans. Further, the effect of predictability
emerges in earlier layers for larger models suggesting that
predictability effects may emerge from greater representational
capacity.

The present results parallel the findings in humans: just as listeners
are slower to recognize \emph{up} within frequent/predictable
V+\emph{up} phrases, a lower classifier logit reflects \emph{up}'s
representation in V+\emph{up} phrases becoming less similar the
representation of the standalone preposition. Under a compositional
account, the representation of \emph{up} should be static regardless of
context. Under usage-based accounts, by contrast, the V+\emph{up} phrase
is compositional at low frequency/predictability, but becomes
progressively less compositional as frequency/predictability increases.
The data supports this graded account specifically: the classifier's
confidence that \emph{up} resembles its prepositional form is highest
for low-frequency/predictability items and decreases as
frequency/predictability rise, rather than staying uniform across all
V+\emph{up} items, as we would expect if the mere presence of a
V+\emph{up} context drove the change. This pattern also cannot be
attributed to the tokenization method: in LLMs, the token for \emph{up}
is identical across all items, whereas each \emph{up} token in Whisper's
audio input is acoustically distinct. This therefore presents a
mechanistic explanation for the slowdown seen in the human behavioral
data.

Lastly, the results suggest that one need not posit separate storage and
abstraction mechanisms to account for the human behavioral data: if
holistic storage and grammatical abstraction arise from the same
process, the traditional distinction between stored items and productive
rules may be a post-hoc description of a continuous representational
landscape rather than a reflection of distinct cognitive processes. The
gradient by-layer results support this view: rather than a sharp
transition from compositional to holistic representations, we observe a
progressive divergence that deepens across layers and strengthens with
frequency and predictability, more consistent with storage-like behavior
being one end of a continuum than with a strict division between stored
and computed items.

Overall, the present study demonstrates another dimension along which
the behavior of transformer models, across modalities, resembles that of
humans, providing further evidence for usage-based theories:
storage-like representations emerge gradually as a function of frequency
and predictability, arising even in models with no explicit storage
mechanism and, for the BabyLMs, no more data than a human receives. More
broadly, the results suggest that transformer architectures offer a
productive framework for investigating not just whether human-like
storage patterns emerge, but \emph{how} and \emph{when} they do so, at a
level of mechanistic detail that behavioral data alone cannot provide.

\section{Limitations}\label{limitations}

The primary limitation is that we examined only one construction
(V+\emph{up} phrases) in one language (English). We chose this
construction deliberately: V+\emph{up} phrasal verbs are among the most
frequent and productive multi-word units in English, and they are the
same construction examined in the human psycholinguistic literature this
study builds on
\citep{kapatsinski2009frequency, houghton2025multiwordrepresentationsminds},
which keeps the model-human comparison as direct as possible. The
tradeoff is depth rather than breadth: rather than testing many
constructions in a single model, we examined this one construction as
thoroughly as our resources allowed, across three model families, two
training modalities (text and speech), models trained on both
human-scale and web-scale data, models varying in number of parameters,
and every hidden layer rather than only the final one. Whether the same
pattern holds for other multi-word constructions or other languages
remains an open question, and one we see as a natural next step rather
than a gap that undermines the present findings. We also only examined
representations at the final checkpoint; however, we look forward to
examining the emergence of storage as a function of training dynamics in
the future.

\bibliography{references}
\clearpage

\onecolumn
% Prevent LaTeX from stretching interparagraph spacing to force pages flush
% to the bottom margin (visible as uneven gaps between paragraphs whenever a
% float relocates away from its declaration point, e.g. Appendix H's Results).
\raggedbottom
\appendix
% Appendix figures may float "here" as well as top/bottom/page (main-text
% figures stay restricted to tbp via the \floatplacement call in the preamble).
\floatplacement{figure}{htbp}

\section{BabyLM Model Training}\label{babylm-model-training}

Models were trained from random initialization for 20 epochs using the
AdamW optimizer with fused weight updates and bfloat16 mixed precision.
The learning rate was set to \(3 \times 10^{-4}\) for the 125M model and
\(1 \times 10^{-4}\) for the 350M and 1.3B models, each preceded by a
linear warmup over 10\% of total training steps. Training was
distributed across two NVIDIA A100 80GB GPUs. Hyperparameter details are
given in Table~\ref{tbl-hyperparams}.

\begin{table}[H]

\centering{

\begin{tabular}{llll}
\toprule
Parameter & OPT-125M & OPT-350M & OPT-1.3B\\
\midrule
Hidden size & 768 & 1,024 & 2,048\\
Attention heads & 12 & 16 & 32\\
Layers & 12 & 24 & 24\\
FFN intermediate size & 3,072 & 4,096 & 8,192\\
Vocabulary size & 8,192 & 8,192 & 8,192\\
Context length (tokens) & 1,024 & 1,024 & 1,024\\
Batch size (per device) & 400 & 50 & 100\\
Gradient accumulation & 1 & 2 & 1\\
Learning rate & $3\times10^{-4}$ & $1.5\times10^{-4}$ & $1\times10^{-4}$\\
Warmup steps & 366 & 732 & 1,465\\
Training epochs & 20 & 20 & 20\\
Optimizer & AdamW (fused) & AdamW (fused) & AdamW (fused)\\
Precision & bfloat16 & bfloat16 & bfloat16\\
\bottomrule
\end{tabular}

}

\caption{\label{tbl-hyperparams}Hyperparameters for the three BabyLM
models. All models share a BPE tokenizer (vocabulary size 8,192) trained
on the BabyLM corpus. Training used NVIDIA A100 80GB GPUs with
distributed data parallelism: two for the 125M and 1.3B, four for the
350M.}

\end{table}%

\clearpage

\section{BabyLM Model Convergence}\label{babylm-model-convergence}

\begin{figure}

\centering{

\pandocbounded{\includegraphics[keepaspectratio]{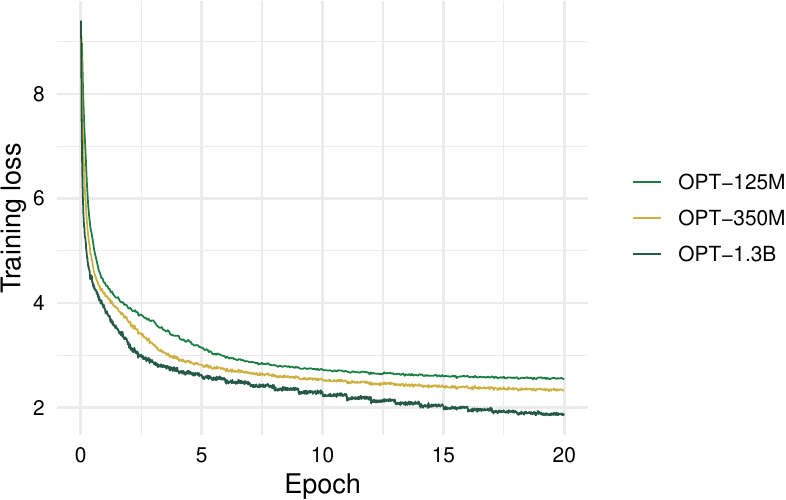}}

}

\caption{\label{fig-train-loss}Training loss over 20 epochs for each
BabyLM model. Loss is logged every 10 gradient steps.}

\end{figure}

All three models trained to completion over 20 epochs, with training
loss decreasing monotonically throughout (Figure~\ref{fig-train-loss}).
Final gradient norms were small for all models (0.16, 0.44, and 0.38 for
the 125M, 350M, and 1.3B models, respectively), and the learning rate
decayed to zero at the end of training, consistent with full convergence
under the scheduled warmup-then-decay regime. No signs of loss
divergence or instability were observed.

The 1.3B model achieves the lowest training loss (1.86), as expected
given its greater capacity, while the 350M model achieves the lowest
validation perplexity (14.35) (Table~\ref{tbl-val-perplexity}).

\begin{table}[H]

\centering{

\begin{tabular}{lrr}
\toprule
Model & Val. loss & Perplexity\\
\midrule
OPT-125M & 2.82 & 16.81\\
OPT-350M & 2.66 & 14.35\\
OPT-1.3B & 2.72 & 15.18\\
\bottomrule
\end{tabular}

}

\caption{\label{tbl-val-perplexity}Final validation loss and perplexity
for each BabyLM model, evaluated on the BabyLM development set after 20
epochs of training.}

\end{table}%

\clearpage

\section{Experiment 1: UP
Independently}\label{experiment-1-up-independently-1}

Validation counts fall slightly below 1,000 for some models because the
tokenizer does not always produce an isolated \emph{up} token; for
example, when \emph{up} appears before punctuation and the tokenizer
fuses the two into a single unit (e.g., \emph{up,}), there is no
isolable \emph{up} position to extract, so that instance is excluded.

\begin{table}[H]

\centering{

\begin{tabular}{lrrrr}
\toprule
Model & Train pos & Train neg & Val pos & Val neg\\
\midrule
OLMo-3 7B & 1,000 & 1,000 & 1,000 & 1,000\\
BabyLM 125M & 1,000 & 1,000 & 999 & 999\\
BabyLM 350M & 1,000 & 1,000 & 999 & 999\\
BabyLM 1.3B & 1,000 & 1,000 & 999 & 999\\
\bottomrule
\end{tabular}

}

\caption{\label{tbl-exp1-splits}Classifier training and validation split
sizes for Experiment 1 (UP independently). Positive (pos) = preposition
\emph{up}; negative (neg) = other alphabetic token from the same
sentence.}

\end{table}%

\begin{table}[H]

\centering{

\begin{tabular}{llrrlr}
\toprule
Model & Coef & Est. & Err. & 95\% CI & \% > 0\\
\midrule
OLMo-3 7B & Intercept & 8.25 & 0.05 & {}[8.15, 8.36] & 100.0\\
OLMo-3 7B & Freq & -1.00 & 0.06 & {}[-1.11, -0.88] & 0.0\\
OLMo-3 7B & Predic. & -2.72 & 0.05 & {}[-2.83, -2.62] & 0.0\\
OLMo-3 7B & Freq:Predic. & -0.01 & 0.05 & {}[-0.12, 0.10] & 41.9\\
BabyLM 125M & Intercept & 7.13 & 0.04 & {}[7.06, 7.20] & 100.0\\
BabyLM 125M & Freq & -0.42 & 0.04 & {}[-0.50, -0.35] & 0.0\\
BabyLM 125M & Predic. & -0.20 & 0.04 & {}[-0.27, -0.12] & 0.0\\
BabyLM 125M & Freq:Predic. & -0.13 & 0.04 & {}[-0.20, -0.05] & 0.0\\
BabyLM 350M & Intercept & 6.46 & 0.05 & {}[6.36, 6.55] & 100.0\\
BabyLM 350M & Freq & -0.71 & 0.05 & {}[-0.81, -0.62] & 0.0\\
BabyLM 350M & Predic. & -0.23 & 0.05 & {}[-0.32, -0.13] & 0.0\\
BabyLM 350M & Freq:Predic. & -0.12 & 0.05 & {}[-0.22, -0.03] & 0.6\\
BabyLM 1.3B & Intercept & 8.21 & 0.05 & {}[8.11, 8.31] & 100.0\\
BabyLM 1.3B & Freq & -0.74 & 0.05 & {}[-0.84, -0.63] & 0.0\\
BabyLM 1.3B & Predic. & -0.14 & 0.05 & {}[-0.24, -0.03] & 0.5\\
BabyLM 1.3B & Freq:Predic. & -0.12 & 0.05 & {}[-0.22, -0.01] & 1.4\\
\bottomrule
\end{tabular}

}

\caption{\label{tbl-exp1-joint-final}Joint frequency × predictability
model at the final layer for UP independently. Estimates are from
Bayesian linear mixed-effects models (brms); \% \textgreater{} 0
indicates the percentage of posterior samples with a positive estimate.}

\end{table}%

\begin{table}[H]

\centering{

\begin{tabular}{llrrl}
\toprule
Model & Predictor & EDF & F & p\\
\midrule
OLMo-3 7B & Frequency & 23.33 & 39583.62 & < 0.001\\
BabyLM 125M & Frequency & 15.32 & 16281.02 & < 0.001\\
BabyLM 350M & Frequency & 21.02 & 39055.63 & < 0.001\\
BabyLM 1.3B & Frequency & 16.76 & 29336.45 & < 0.001\\
OLMo-3 7B & Predictability & 23.40 & 45016.89 & < 0.001\\
BabyLM 125M & Predictability & 22.00 & 12479.55 & < 0.001\\
BabyLM 350M & Predictability & 22.56 & 35627.86 & < 0.001\\
BabyLM 1.3B & Predictability & 22.39 & 23854.74 & < 0.001\\
\bottomrule
\end{tabular}

}

\caption{\label{tbl-exp1-gam-smooth}GAM tensor-product smooth summary
for UP independently. EDF = estimated degrees of freedom; p = p-value
for the te(predictor, layer) smooth term.}

\end{table}%

\begin{figure}

\centering{

\pandocbounded{\includegraphics[keepaspectratio]{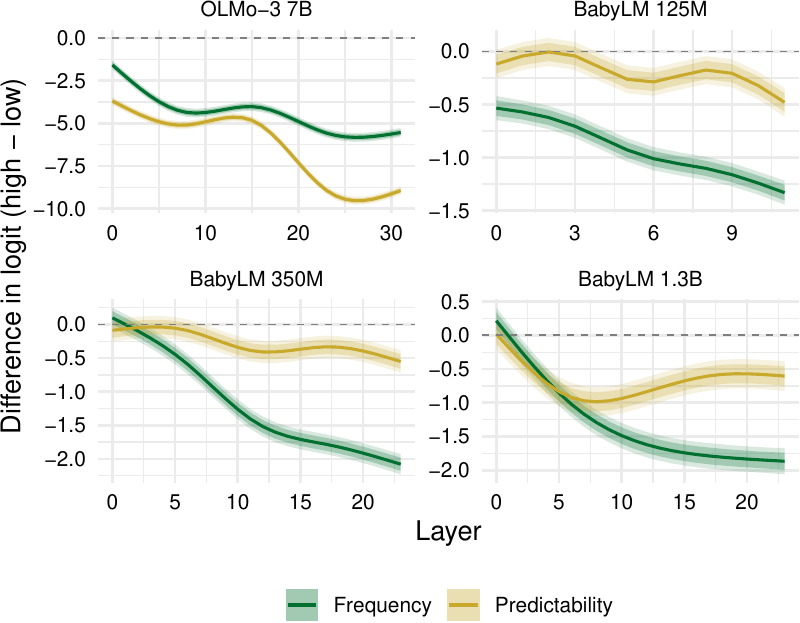}}

}

\caption{\label{fig-exp1-gam-diffs}Layer-by-layer difference in
GAM-predicted logit between high (90th percentile) and low (10th
percentile) predictor values, for UP independently. Green = frequency
effect; yellow = predictability effect. Inner and outer shaded ribbons
show 80\% and 95\% confidence intervals derived from the GAM standard
error. A negative difference indicates that high-predictor phrases yield
a lower classifier logit (i.e., their representation of \emph{up} is
less similar to the prepositional class).}

\end{figure}%

\clearpage

\section{Classifier Validation Accuracy by
Layer}\label{app-val-accuracy}

Tables \ref{xtblval-acc-olmo}--\ref{xtblval-acc-whisper} report held-out
validation accuracy for every classifier trained in this study, broken
down by layer, model, and condition (Experiment 1: independent
\emph{up}; Experiment 2: \emph{up} as a subword within a larger word;
Experiment 3: Whisper encoder/decoder).

\begin{longtable}{rrr}
\caption{\label{xtblval-acc-olmo}OLMo-3 7B validation accuracy (\%) by
layer, Experiment 1 (independent \emph{up}) vs.\ Experiment 2 (\emph{up}
as subword).}\\
\toprule
Layer & Exp.\ 1 (Indep.) & Exp.\ 2 (Subword)\\
\midrule
\endfirsthead
\toprule
Layer & Exp.\ 1 (Indep.) & Exp.\ 2 (Subword)\\
\midrule
\endhead
\bottomrule
\endfoot
0 & 100.0 & 99.7 \\
1 & 100.0 & 99.7 \\
2 & 100.0 & 99.7 \\
3 & 100.0 & 99.7 \\
4 & 99.9 & 99.5 \\
5 & 100.0 & 99.5 \\
6 & 99.9 & 99.5 \\
7 & 100.0 & 99.3 \\
8 & 99.9 & 99.2 \\
9 & 100.0 & 99.1 \\
10 & 100.0 & 98.9 \\
11 & 100.0 & 98.9 \\
12 & 100.0 & 98.6 \\
13 & 100.0 & 98.9 \\
14 & 100.0 & 98.5 \\
15 & 100.0 & 98.6 \\
16 & 100.0 & 98.6 \\
17 & 99.9 & 98.6 \\
18 & 100.0 & 98.5 \\
19 & 100.0 & 98.4 \\
20 & 99.9 & 98.4 \\
21 & 99.9 & 98.3 \\
22 & 99.8 & 98.3 \\
23 & 99.8 & 98.3 \\
24 & 99.7 & 98.0 \\
25 & 99.6 & 98.1 \\
26 & 99.6 & 97.8 \\
27 & 99.6 & 97.9 \\
28 & 99.6 & 97.5 \\
29 & 99.6 & 97.6 \\
30 & 99.3 & 97.2 \\
31 & 99.4 & 97.4 \\
\end{longtable}

\begin{longtable}{rrrrrrr}
\caption{\label{xtblval-acc-babylm}BabyLM validation accuracy (\%) by
layer and model, Experiment 1 (independent \emph{up}) vs.\ Experiment 2
(\emph{up} as subword). OPT-125M has 12 layers; ``--'' marks layers
beyond a model's depth.}\\
\toprule
Layer & \multicolumn{2}{c}{OPT-125M} & \multicolumn{2}{c}{OPT-350M} & \multicolumn{2}{c}{OPT-1.3B}\\
 & Exp.\ 1 & Exp.\ 2 & Exp.\ 1 & Exp.\ 2 & Exp.\ 1 & Exp.\ 2\\
\midrule
\endfirsthead
\toprule
Layer & \multicolumn{2}{c}{OPT-125M} & \multicolumn{2}{c}{OPT-350M} & \multicolumn{2}{c}{OPT-1.3B}\\
 & Exp.\ 1 & Exp.\ 2 & Exp.\ 1 & Exp.\ 2 & Exp.\ 1 & Exp.\ 2\\
\midrule
\endhead
\bottomrule
\endfoot
0 & 100.0 & 99.3 & 100.0 & 99.5 & 100.0 & 99.2 \\
1 & 100.0 & 99.2 & 99.9 & 99.7 & 100.0 & 99.3 \\
2 & 100.0 & 99.2 & 99.9 & 99.5 & 100.0 & 99.2 \\
3 & 100.0 & 99.2 & 100.0 & 99.3 & 100.0 & 99.1 \\
4 & 100.0 & 99.1 & 100.0 & 99.4 & 100.0 & 99.0 \\
5 & 100.0 & 98.9 & 100.0 & 99.5 & 100.0 & 99.2 \\
6 & 100.0 & 99.1 & 100.0 & 99.5 & 100.0 & 99.1 \\
7 & 100.0 & 99.1 & 100.0 & 99.4 & 100.0 & 99.1 \\
8 & 100.0 & 99.0 & 100.0 & 99.3 & 100.0 & 99.1 \\
9 & 100.0 & 99.0 & 100.0 & 99.5 & 100.0 & 99.1 \\
10 & 100.0 & 98.8 & 100.0 & 99.5 & 100.0 & 99.2 \\
11 & 100.0 & 98.6 & 100.0 & 99.5 & 100.0 & 99.2 \\
12 & -- & -- & 100.0 & 99.6 & 100.0 & 99.1 \\
13 & -- & -- & 100.0 & 99.7 & 100.0 & 99.3 \\
14 & -- & -- & 100.0 & 99.7 & 100.0 & 99.2 \\
15 & -- & -- & 100.0 & 99.7 & 100.0 & 99.0 \\
16 & -- & -- & 100.0 & 99.6 & 100.0 & 99.0 \\
17 & -- & -- & 100.0 & 99.6 & 100.0 & 98.9 \\
18 & -- & -- & 100.0 & 99.5 & 100.0 & 99.2 \\
19 & -- & -- & 99.9 & 99.2 & 100.0 & 99.1 \\
20 & -- & -- & 99.9 & 99.3 & 99.9 & 99.1 \\
21 & -- & -- & 99.9 & 99.1 & 99.9 & 99.0 \\
22 & -- & -- & 99.9 & 99.0 & 100.0 & 99.1 \\
23 & -- & -- & 99.9 & 99.1 & 100.0 & 99.1 \\
\end{longtable}

\begin{longtable}{rrr}
\caption{\label{xtblval-acc-whisper}Whisper validation accuracy (\%) by
layer and component, Experiment 3 (independent \emph{up}). The Whisper
subword replication (a robustness check, not a numbered experiment) is
reported separately in Table~\ref{xtblexp3-sub-val-acc}
(Appendix~\ref{app-whisper-subword}).}\\
\toprule
Layer & Encoder & Decoder\\
\midrule
\endfirsthead
\toprule
Layer & Encoder & Decoder\\
\midrule
\endhead
\bottomrule
\endfoot
0 & 96.8 & 99.8 \\
1 & 97.5 & 100.0 \\
2 & 98.2 & 100.0 \\
3 & 98.1 & 99.9 \\
4 & 98.5 & 100.0 \\
5 & 98.7 & 100.0 \\
6 & 99.0 & 100.0 \\
7 & 99.1 & 100.0 \\
8 & 99.3 & 99.9 \\
9 & 99.3 & 100.0 \\
10 & 99.5 & 99.7 \\
11 & 99.5 & 99.4 \\
\end{longtable}

\clearpage

\section{Test Set Items}\label{test-set-items}

Frequency and predictability are correlated with one another to some
degree (r = 0.34 for OLMo-3 7B, r = 0.25 for BabyLM, r = 0.31 for
Whisper; see Table~\ref{tbl-test-items} for exact per-model values).
Because both predictors are included in the same model rather than fit
separately, each reported coefficient reflects the effect of that
predictor above and beyond the other; the corresponding variance
inflation factors are low across all models, indicating that this
correlation does not meaningfully compromise the reliability of either
estimate.

\begin{table}[H]

\centering{

\begin{tabular}{lrrlrlrr}
\toprule
Model & N items & Med. freq & Freq range & Med. predic. & Predic. range & $r$ & VIF\\
\midrule
OLMo-3 7B & 4081 & 43608 & 2--103,842,740 & -5.38 & -10.98--2.74 & 0.34 & 1.13\\
BabyLM & 1039 & 34 & 10--9,963 & -3.40 & -10.38--1.99 & 0.25 & 1.07\\
Whisper encoder & 1460 & 235806 & 960--103,842,740 & -4.08 & -10.98--2.74 & 0.31 & 1.10\\
Whisper decoder & 1460 & 235806 & 960--103,842,740 & -4.08 & -10.98--2.74 & 0.31 & 1.10\\
\bottomrule
\end{tabular}

}

\caption{\label{tbl-test-items}Test set statistics for all experiments.
\(r\) and VIF give the Pearson correlation and variance inflation factor
between log-frequency and log-predictability for that model's test
items.}

\end{table}%

\clearpage

\section{Experiment 2: UP as
Subword}\label{experiment-2-up-as-subword-1}

\begin{table}[H]

\centering{

\begin{tabular}{lrrrr}
\toprule
Model & Train pos & Train neg & Val pos & Val neg\\
\midrule
OLMo-3 7B & 2,000 (1,000) & 2,000 & 1,876 (876) & 1,876\\
BabyLM 125M & 2,000 (1,000) & 2,000 & 1,582 (583) & 1,582\\
BabyLM 350M & 2,000 (1,000) & 2,000 & 1,582 (583) & 1,582\\
BabyLM 1.3B & 2,000 (1,000) & 2,000 & 1,582 (583) & 1,582\\
\bottomrule
\end{tabular}

}

\caption{\label{tbl-exp2-splits}Classifier training and validation split
sizes for Experiment 2 (UP as subword). Positive (pos) examples combine
1,000 prepositional \emph{up} instances with up to 1,000 unique
up-within-word types; parenthetical counts give the subword-specific
portion (below 1,000 in validation where not every sampled instance
yielded a resolvable token position). Negative (neg) = other alphabetic
token from the same sentence pool. Val pos and Val neg are the balanced
(truncated to equal class size) counts actually used for evaluation, as
in Tables \ref{tbl-exp1-splits} and \ref{tbl-exp3-splits}.}

\end{table}%

\begin{table}[H]

\centering{

\begin{tabular}{llrrlr}
\toprule
Model & Coef & Est. & Err. & 95\% CI & \% > 0\\
\midrule
OLMo-3 7B & Intercept & 10.75 & 0.05 & {}[10.64, 10.84] & 100.0\\
OLMo-3 7B & Freq & -0.84 & 0.05 & {}[-0.95, -0.74] & 0.0\\
OLMo-3 7B & Predic. & -2.00 & 0.05 & {}[-2.10, -1.91] & 0.0\\
OLMo-3 7B & Freq:Predic. & -0.05 & 0.05 & {}[-0.15, 0.05] & 14.6\\
BabyLM 125M & Intercept & 11.63 & 0.08 & {}[11.47, 11.79] & 100.0\\
BabyLM 125M & Freq & -0.76 & 0.08 & {}[-0.92, -0.59] & 0.0\\
BabyLM 125M & Predic. & -0.09 & 0.08 & {}[-0.25, 0.07] & 13.0\\
BabyLM 125M & Freq:Predic. & -0.30 & 0.08 & {}[-0.46, -0.13] & 0.0\\
BabyLM 350M & Intercept & 7.01 & 0.09 & {}[6.84, 7.19] & 100.0\\
BabyLM 350M & Freq & -0.71 & 0.09 & {}[-0.89, -0.53] & 0.0\\
BabyLM 350M & Predic. & -0.87 & 0.09 & {}[-1.05, -0.68] & 0.0\\
BabyLM 350M & Freq:Predic. & -0.19 & 0.09 & {}[-0.37, -0.01] & 1.7\\
BabyLM 1.3B & Intercept & 10.15 & 0.05 & {}[10.04, 10.25] & 100.0\\
BabyLM 1.3B & Freq & -0.84 & 0.05 & {}[-0.95, -0.74] & 0.0\\
BabyLM 1.3B & Predic. & 0.14 & 0.05 & {}[0.04, 0.25] & 99.6\\
BabyLM 1.3B & Freq:Predic. & -0.03 & 0.05 & {}[-0.13, 0.08] & 30.4\\
\bottomrule
\end{tabular}

}

\caption{\label{tbl-exp2-joint-final}Joint frequency × predictability
model at the final layer for UP as subword. Estimates are from Bayesian
linear mixed-effects models (brms); \% \textgreater{} 0 indicates the
percentage of posterior samples with a positive estimate.}

\end{table}%

\begin{table}[H]

\centering{

\begin{tabular}{llrrl}
\toprule
Model & Predictor & EDF & F & p\\
\midrule
OLMo-3 7B & Frequency & 23.36 & 44489.41 & < 0.001\\
BabyLM 125M & Frequency & 16.98 & 485.69 & < 0.001\\
BabyLM 350M & Frequency & 21.10 & 9188.77 & < 0.001\\
BabyLM 1.3B & Frequency & 19.74 & 16171.80 & < 0.001\\
OLMo-3 7B & Predictability & 23.12 & 48483.14 & < 0.001\\
BabyLM 125M & Predictability & 20.52 & 331.07 & < 0.001\\
BabyLM 350M & Predictability & 22.20 & 8818.43 & < 0.001\\
BabyLM 1.3B & Predictability & 21.83 & 15067.89 & < 0.001\\
\bottomrule
\end{tabular}

}

\caption{\label{tbl-exp2-gam-smooth}GAM tensor-product smooth summary
for UP as subword. EDF = estimated degrees of freedom; p = p-value for
the te(predictor, layer) smooth term.}

\end{table}%

\begin{figure}

\centering{

\pandocbounded{\includegraphics[keepaspectratio]{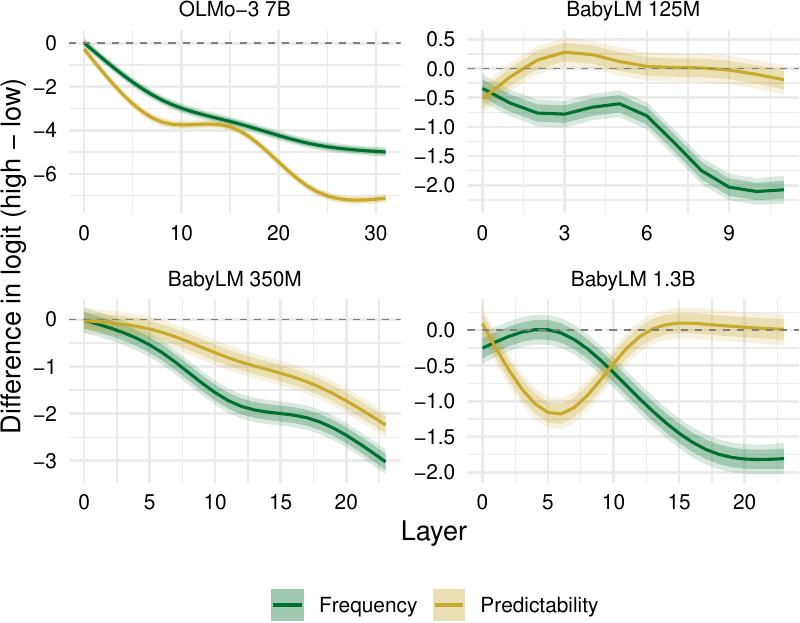}}

}

\caption{\label{fig-exp2-gam-diffs}Layer-by-layer difference in
GAM-predicted logit between high (90th percentile) and low (10th
percentile) predictor values. Green = frequency effect; yellow =
predictability effect. Inner and outer shaded ribbons show 80\% and 95\%
confidence intervals derived from the GAM standard error. A negative
difference indicates that high-predictor phrases yield a lower
classifier logit (i.e., their representation of \emph{up} is less
similar to the preposition class).}

\end{figure}%

\clearpage

\section{Whisper Subword Replication}\label{app-whisper-subword}

As a further robustness check, we replicated Experiment 2's subword
design for Whisper, applying the same up-within-word criteria used there
but adapted for audio.

\subsection{Methods}\label{app-whisper-subword-methods}

\subsubsection{Classifier
Training}\label{app-whisper-subword-classifier-training}

The classifier's positive training class was broadened to include
instances of \emph{up} embedded within a larger word in addition to the
standalone (prepositional) \emph{up}. Unlike Experiment 2's purely
orthographic candidate selection, audio candidates were additionally
required to actually be \emph{pronounced} with the \emph{up} sound: each
candidate word's CMU Pronouncing Dictionary transcription was checked
for a contiguous AH-P phoneme pair (covering both the stressed STRUT
vowel, as in \emph{couple}, and the unstressed schwa, as in
\emph{support}), which excludes words that merely spell \emph{up}
without saying it (e.g., \emph{superintendent}). Since audio has no
discrete subword tokens, character-level WhisperX forced alignment was
then used to isolate the \emph{up} portion of each host word's audio
span.

Unlike Experiment 2's text-side design, which draws exactly one instance
per unique up-within-word type (a design that works there because
roughly 8,662 qualifying types are available in the text corpora), the
audio corpus yields far fewer qualifying types, even with a lower floor
for inclusion: only a few hundred up-word types meet this bar.

Up-word types were therefore first split into disjoint train and
validation sets, so that no type appears in both, and only then were up
to five audio instances sampled per type within its assigned split,
rather than exactly one. Allowing multiple instances per type is a
smaller concession to memorization risk here than it would be for text:
because each instance is drawn from a different underlying audio segment
and speaker, repeated instances of the same up-word type still give the
classifier acoustically distinct embeddings rather than the same feature
vector encountered multiple times, unlike, for instance, a token
embedding that would recur identically across repeated occurrences of
the same word. Negative examples paired with these positives were drawn
from the same audio segment and excluded not just the exact word
\emph{up} but any word containing \emph{up} as a substring, so a
negative could not accidentally be a second up-containing word from the
same sentence. Because types themselves are still held disjoint between
train and validation, validation accuracy on the subword condition
specifically reflects the classifier's performance on up-within-word
types it never encountered during training, rather than repeated
exposure to previously-seen words, and is reported separately from
validation accuracy on prepositional \emph{up}
(Table~\ref{xtblexp3-sub-val-acc}). The test set is otherwise identical
to Experiment 3.

\subsubsection{Analyses}\label{app-whisper-subword-analyses}

Predictor definitions are otherwise identical to Experiment 3. Following
Experiment 3's own design, both the joint model at the final layer and
the by-layer GAM analyses additionally include the duration of the
\emph{up} segment (\(c\_\text{duration}\)) as a covariate, to control
for the same phonetic-reduction confound
(Equation~\ref{eq-linear-joint-duration} and
Equation~\ref{eq-gam-layers-duration}, respectively). Unlike the
main-text analyses, we additionally fit single-predictor ``frequency
alone'' and ``predictability alone'' models (likewise controlling for
duration) here, in order to diagnose the joint model's frequency
coefficient patterns in both components, discussed in
Appendix~\ref{app-whisper-subword-suppression} below.

\subsection{Results}\label{app-whisper-subword-results}

Two patterns depart from the results in the main text, for different
reasons. The decoder's joint-model frequency coefficient reverses sign
relative to the rest of the paper; as we demonstrate in
Appendix~\ref{app-whisper-subword-suppression} below, this is a
statistical-suppression artifact of the frequency-predictability
correlation, not evidence that the underlying (negative-leaning)
frequency effect fails to replicate. The coefficient for frequency in
the encoder is also positive here, but for a different reason: it is
credible both in the joint model and in the single-predictor model alone
(Table~\ref{xtblexp3-sub-alone}), so it is not a suppression artifact
but a genuine, reproducible positive effect, a stronger version of the
same directional, though non-credible, pattern already present in the
main experiment's encoder. The effect of predictability, by contrast,
replicates as negative in both components throughout, exactly as in the
main experiment.

\begin{figure}

\centering{

\pandocbounded{\includegraphics[keepaspectratio]{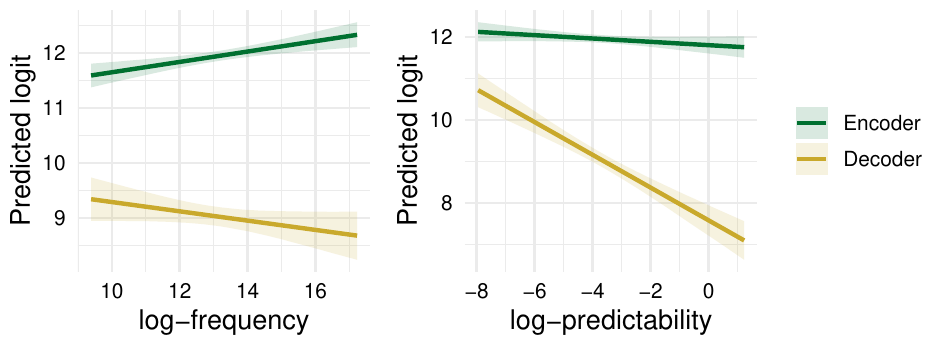}}

}

\caption{\label{xfigexp3-sub-marginal-effects}Predicted logit by
frequency (left) and predictability (right) for Whisper encoder and
decoder, subword replication, from the individual (single-predictor)
models rather than the joint model. Shading indicates 95\% CI.}

\end{figure}

\begin{figure}

\centering{

\pandocbounded{\includegraphics[keepaspectratio]{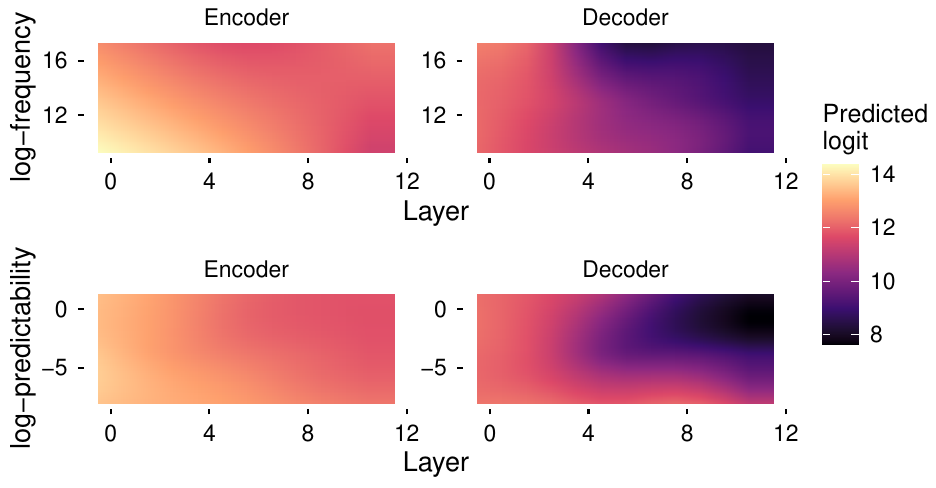}}

}

\caption{\label{xfigexp3-sub-gam-layers}GAM-predicted logit by layer
and predictor for Whisper encoder and decoder, subword replication. Top:
frequency; bottom: predictability. Color = logit; darker = lower logit
(less similar to the preposition \emph{up}); random effects excluded.}

\end{figure}

\begin{figure}

\centering{

\pandocbounded{\includegraphics[keepaspectratio]{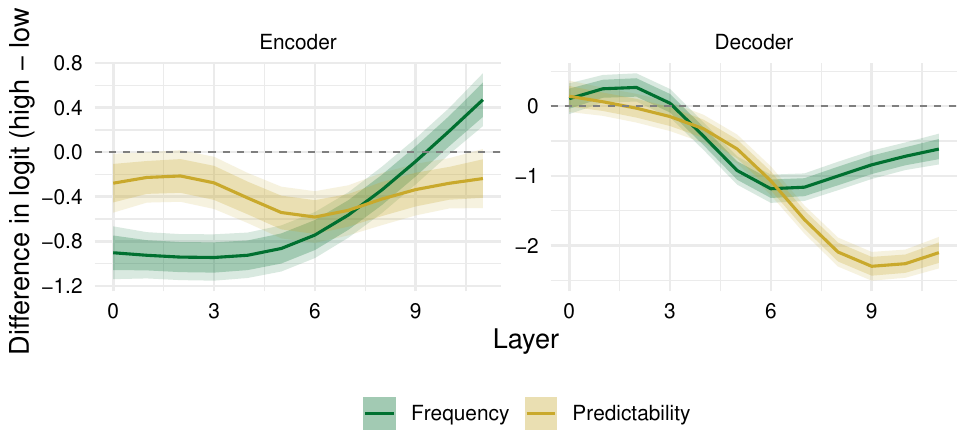}}

}

\caption{\label{xfigexp3-sub-gam-diffs}Layer-by-layer difference in
GAM-predicted logit between high (90th percentile) and low (10th
percentile) predictor values, for Whisper encoder and decoder, subword
replication. Green = frequency effect; yellow = predictability effect.
Inner and outer shaded ribbons show 80\% and 95\% confidence intervals
derived from the GAM standard error. A negative difference indicates
that high-predictor phrases yield a lower classifier logit (i.e., their
representation of \emph{up} is less similar to the standalone, preposition class).}

\end{figure}

Split sizes are given in Table~\ref{xtblexp3-sub-splits}. Per-layer
validation accuracy, broken down into overall (combined
standalone-\emph{up} + subword-\emph{up} positives) and subword-only
accuracy, is given in Table~\ref{xtblexp3-sub-val-acc}. Full numerical
results are reported in Table~\ref{xtblexp3-sub-alone} (single-predictor
models), Table~\ref{xtblexp3-sub-joint-final} (joint model), and
Table~\ref{xtblexp3-sub-gam-smooth} (GAM); the individual
(single-predictor) model effects are visualized in
Figure~\ref{xfigexp3-sub-marginal-effects} and
Figure~\ref{xfigexp3-sub-gam-layers}, with
Figure~\ref{xfigexp3-sub-gam-diffs} showing the layer-by-layer
difference directly.

Considered on their own (Table~\ref{xtblexp3-sub-alone}), the two
predictors show different patterns by component, controlling for
duration as in the joint model. In the decoder, both frequency and
predictability lean negative, matching the pattern found throughout the
paper: predictability is credible, and frequency is negative-leaning
(95.8\% of the posterior below zero) though just short of the
conventional 95\% threshold. It is this negative-leaning frequency
effect, not a genuine positive one, that the joint model's sign flip
(discussed below) obscures. In the encoder, predictability is likewise
negative-leaning (94.5\% below zero) but frequency shows a different
pattern: it is credibly \emph{positive} on its own (100.0\% of the
posterior above zero), matching the sign and credibility of the
joint-model coefficient reported below. Because the alone and joint
models agree here, this cannot be a suppression artifact, unlike the
decoder's frequency coefficient, where alone and joint disagree.

\begin{table}[H]

\centering{

\begin{tabular}{llrrlr}
\toprule
Model & Coef & Est. & Err. & 95\% CI & \% > 0\\
\midrule
Encoder & Freq. (alone) & 0.09 & 0.03 & {}[0.04, 0.15] & 100.0\\
Encoder & Predic. (alone) & -0.04 & 0.03 & {}[-0.09, 0.01] & 5.5\\
Decoder & Freq. (alone) & -0.09 & 0.05 & {}[-0.18, 0.01] & 4.2\\
Decoder & Predic. (alone) & -0.40 & 0.04 & {}[-0.48, -0.31] & 0.0\\
\bottomrule
\end{tabular}

}

\caption{\label{xtblexp3-sub-alone}Single-predictor ``frequency alone'' and
``predictability alone'' models for the Whisper subword replication, each
also controlling for duration ($c\_\text{duration}$) as in the joint
model. Estimates are from Bayesian mixed-effects models (brms); \%
\textgreater{} 0 indicates the percentage of posterior samples with a
positive estimate.}

\end{table}

In the joint model, the coefficient for predictability remained credible
in both components and was, if anything, larger than in the results in
the main text. The coefficient for frequency in the encoder replicates
the main experiment's direction. In the decoder, the effect of frequency
diverges from that of the main experiment, with its CI crossing zero and
its sign flipping to positive. This sign flip is not due to a lack of an
effect, but rather due to statistical suppression, which we demonstrate
in the following section.

\begin{table}[H]

\centering{

\begin{tabular}{lrrrr}
\toprule
Component & Train pos & Train neg & Val pos & Val neg\\
\midrule
Encoder & 1,775 (775) & 1,775 & 1,775 (775) & 1,775\\
Decoder & 1,686 (686) & 1,686 & 1,696 (696) & 1,696\\
\bottomrule
\end{tabular}

}

\caption{\label{xtblexp3-sub-splits}Classifier training and validation
split sizes for the Whisper subword replication. Positive examples
combine 1,000 prepositional \emph{up} instances (as in Experiment 3) with
up to five instances per up-within-word type; parenthetical counts give
the subword-specific portion. Negatives are matched 1:1 using the same
criteria as Experiment 3.}

\end{table}

\begin{table}[H]

\centering{

\begin{tabular}{rrrrr}
\toprule
Layer & \multicolumn{2}{c}{Encoder} & \multicolumn{2}{c}{Decoder}\\
 & Overall & Subword & Overall & Subword\\
\midrule
0 & 94.8 & 91.0 & 97.7 & 90.9 \\
1 & 95.8 & 93.0 & 98.0 & 91.5 \\
2 & 97.0 & 94.8 & 98.2 & 92.7 \\
3 & 97.5 & 96.0 & 98.4 & 93.5 \\
4 & 98.0 & 96.9 & 97.9 & 91.5 \\
5 & 98.5 & 97.4 & 97.8 & 91.2 \\
6 & 98.7 & 97.9 & 97.9 & 92.0 \\
7 & 98.7 & 98.2 & 98.0 & 93.1 \\
8 & 98.8 & 97.9 & 97.7 & 92.2 \\
9 & 99.0 & 97.5 & 97.1 & 90.2 \\
10 & 99.0 & 97.8 & 95.7 & 87.6 \\
11 & 99.2 & 98.2 & 94.5 & 84.1 \\
\bottomrule
\end{tabular}

}

\caption{\label{xtblexp3-sub-val-acc}Held-out validation accuracy (\%) by
layer and component for the Whisper subword replication. Overall =
accuracy on the full combined validation set (standalone-\emph{up} +
subword-\emph{up} positives vs.\ negatives); Subword = accuracy
restricted to the subword-\emph{up} validation rows only, i.e.\
up-within-word types never seen during training (see discussion above).
The main text's aggregate validation-accuracy claim
(Appendix~\ref{app-val-accuracy}) uses the Overall column; the Subword
column is reported here to make the type-generalization check described
above concrete.}

\end{table}

\begin{table}[H]

\centering{

\begin{tabular}{llrrlr}
\toprule
Model & Coef & Est. & Err. & 95\% CI & \% > 0\\
\midrule
Encoder & Intercept & 11.98 & 0.05 & {}[11.88, 12.07] & 100.0\\
Encoder & Freq & 0.19 & 0.05 & {}[0.10, 0.28] & 100.0\\
Encoder & Predic. & -0.14 & 0.05 & {}[-0.23, -0.04] & 0.3\\
Encoder & Duration & -0.10 & 0.03 & {}[-0.16, -0.05] & 0.0\\
Encoder & Freq:Predic. & -0.03 & 0.05 & {}[-0.11, 0.06] & 28.2\\
Decoder & Intercept & 9.05 & 0.09 & {}[8.88, 9.21] & 100.0\\
Decoder & Freq & 0.05 & 0.08 & {}[-0.11, 0.22] & 74.1\\
Decoder & Predic. & -0.78 & 0.09 & {}[-0.95, -0.60] & 0.0\\
Decoder & Duration & 0.08 & 0.04 & {}[-0.01, 0.17] & 96.7\\
Decoder & Freq:Predic. & -0.12 & 0.08 & {}[-0.28, 0.04] & 6.8\\
\bottomrule
\end{tabular}

}

\caption{\label{xtblexp3-sub-joint-final}Joint frequency × predictability
model at the final layer for the Whisper subword replication, controlling
for the duration of the \emph{up} audio segment (Equation~\ref{eq-linear-joint-duration}).
Estimates are from Bayesian linear mixed-effects models (brms); \% \textgreater{}
0 indicates the percentage of posterior samples with a positive
estimate.}

\end{table}

\begin{table}[H]

\centering{

\begin{tabular}{llrrl}
\toprule
Model & Predictor & EDF & F & p\\
\midrule
Encoder & Frequency & 7.38 & 289.17 & < 0.001\\
Decoder & Frequency & 20.42 & 1144.50 & < 0.001\\
Encoder & Predictability & 13.43 & 132.55 & < 0.001\\
Decoder & Predictability & 21.79 & 1184.45 & < 0.001\\
\bottomrule
\end{tabular}

}

\caption{\label{xtblexp3-sub-gam-smooth}GAM tensor-product smooth
summary for the Whisper subword replication, encoder and decoder,
controlling for the duration of the \emph{up} audio segment
(Equation~\ref{eq-gam-layers-duration}), matching Experiment 3's own
by-layer GAM design. EDF = estimated degrees of freedom; p = p-value
for the te(predictor, layer) smooth term.}

\end{table}

\subsection{Statistical
Suppression}\label{app-whisper-subword-suppression}

Statistical suppression occurs when two correlated predictors, entered
into the same regression, produce a coefficient for one of them that
differs in sign from its own independent relationship with the outcome,
even though neither predictor's independent relationship with the
outcome is itself in question
\citep{friedman2005graphicalviewssuppression}. This is not a sign of an
unreliable model: the standard collinearity diagnostic, the variance
inflation factor (VIF), is 1.10 for this predictor pairing, comparable
to every other analysis in this paper (1.07--1.13 across all model
families). VIF depends only on how correlated the two predictors are
with each other, not on how strongly each is independently related to
the outcome, so it cannot itself flag which specific analyses are at
risk of this pattern.

When two predictors are entered into the same standardized regression,
the coefficient of whichever predictor has the smaller correlation with
the outcome is:

\begin{equation}\phantomsection\label{eq-suppression-partial}{
\hat\beta_{\text{weak}} = \frac{r_{y,\text{weak}} - r_{y,\text{strong}} \cdot r_{1,2}}{1 - r_{1,2}^2}
}\end{equation}

where \(r_{y,\text{weak}}\) and \(r_{y,\text{strong}}\) are the two
predictors' zero-order correlations with the outcome and \(r_{1,2}\) is
the correlation between the predictors themselves
\citep{friedman2005graphicalviewssuppression}. The denominator is always
positive, so the sign of \(\hat\beta_{\text{weak}}\) depends entirely on
the numerator, which turns negative exactly when:

\begin{equation}\phantomsection\label{eq-suppression-threshold}{
r_{1,2} > \frac{r_{y,\text{weak}}}{r_{y,\text{strong}}}
}\end{equation}

Table~\ref{xtblexp3-sub-suppression} computes both sides of this
inequality for every model and condition in the paper. The predictor
correlation \(r_{1,2}\) is fixed by the shared test set and is identical
across conditions for a given model. Three of the ten conditions have an
observed \(r_{1,2}\) exceeding this threshold. Everywhere else, the same
background correlation stays below each condition's threshold, so those
coefficients can only attenuate, never reverse.

Exceeding the threshold, however, only flags a condition as \emph{at
risk} of suppression; it cannot by itself distinguish a real effect
being suppressed from a case where the weaker predictor's own
relationship is simply too close to zero for the linear approximation to
be informative either way. We adjudicate this directly using the
single-predictor (alone) models, and the three flagged Whisper
conditions turn out to fall into three different categories. In the
subword encoder, the alone-model frequency coefficient is credibly
\emph{positive} (Table~\ref{xtblexp3-sub-alone}), agreeing with the
joint model rather than contradicting it; there is nothing to suppress,
and we treat this as a genuine, replicating positive effect. In the
indep encoder, by contrast, neither the alone-model estimate (81.4\% of
the posterior above zero) nor the joint-model estimate (91.0\%,
Table~\ref{tbl-exp3-joint-final}) reaches credibility in either
direction; this is not suppression either, since there is no credible
effect on either side to be masked or revealed, so we treat frequency in
the indep encoder as not meaningfully different from zero. Only the
decoder-subword condition shows the alone/joint disagreement that
suppression specifically predicts: frequency when modeled without
predictability is negative-leaning (95.8\% of the posterior below zero,
matching frequency's sign in every other condition in the paper), while
in the joint model it is non-credible and positive-leaning
(Table~\ref{xtblexp3-sub-alone}, Table~\ref{xtblexp3-sub-joint-final}).
We therefore reserve the suppression explanation for the decoder-subword
condition alone.

\begin{table}[H]

\centering{

\begin{tabular}{lrrrrl}
\toprule
Condition & $r(Y,\text{freq})$ & $r(Y,\text{predic})$ & $r_{1,2}$ & Threshold & Result\\
\midrule
OLMo indep & -0.338 & -0.539 & 0.337 & 0.628 & stable\\
OLMo subword & -0.294 & -0.438 & 0.337 & 0.672 & stable\\
BabyLM-125M indep & -0.287 & -0.159 & 0.251 & 0.554 & stable\\
BabyLM-125M subword & -0.212 & -0.057 & 0.251 & 0.267 & stable\\
BabyLM-350M indep & -0.297 & -0.100 & 0.251 & 0.338 & stable\\
BabyLM-1.3B indep & -0.316 & -0.119 & 0.251 & 0.375 & stable\\
Whisper encoder indep & 0.013 & -0.017 & 0.306 & -0.748 & sign flip\\
Whisper encoder subword & 0.039 & -0.018 & 0.306 & -0.465 & sign flip\\
Whisper decoder indep & -0.079 & -0.193 & 0.306 & 0.411 & stable\\
Whisper decoder subword & -0.022 & -0.129 & 0.306 & 0.168 & sign flip\\
\bottomrule
\end{tabular}

}

\caption{\label{xtblexp3-sub-suppression}Threshold is the right-hand side of
Equation~\ref{eq-suppression-threshold} for that condition; Result
indicates whether the observed $r_{1,2}$ exceeds it. Two BabyLM subword
conditions (350M, 1.3B) are omitted: predictability's zero-order
correlation with the logit is already near zero there prior to any joint
modelling, a distinct phenomenon not addressed by this mechanism.}

\end{table}

Equation~\ref{eq-suppression-threshold} is derived for a simple
two-predictor regression on standardized variables; our fitted models
additionally include a frequency-by-predictability interaction and a
random intercept for verb, and are estimated in a Bayesian rather than
least-squares framework. Recomputing the decoder's frequency coefficient
with each of these added incrementally shows the sign is unaffected by
any of them: the coefficient remains positive in the subword condition
and negative in the indep condition at every stage, though its magnitude
shifts somewhat under the random-intercept specification. The threshold
values in Table~\ref{xtblexp3-sub-suppression} should therefore be read
as illustrating the mechanism rather than as an exact algebraic
derivation of the reported posterior estimates.

An alternative explanation is that frequency's negative zero-order
correlation is itself an artifact of its correlation with
predictability, and the joint model reveals frequency's true,
non-negative relationship rather than suppressing a real one.
Suppression requires two things at once: the predictors must be
correlated, and the weaker predictor's own relationship to the outcome
must be small enough for that correlation to swamp it. The first
condition holds equally across all ten conditions in this paper
(\(r_{1,2} = 0.251\)--\(0.337\)), so it cannot by itself explain why any
particular condition reverses. Comparing the magnitude of frequency's
zero-order correlation across conditions is not decisive here either,
since all three flagged Whisper conditions have small values (\(-0.022\)
to \(0.039\)) and yet only one is genuine suppression, as established
directly above via the alone-model comparison. That comparison, rather
than the magnitude of the zero-order correlation alone, is what
distinguishes the decoder-subword condition (alone and joint disagree)
from the two encoder conditions (alone and joint agree, whether both
credible or both non-credible).

Taken together: predictability replicates as a negative effect
throughout; the decoder's frequency departure has a specific,
well-understood statistical explanation (suppression) rather than
reflecting a substantive non-replication; and the encoder's frequency
effect is not a departure from the main experiment at all, but a
credible version of the same small positive lean already present there.
We therefore consider the subword replication to parallel Experiment 3's
results throughout, including in the decoder.

\section{Experiment 3: Whisper}\label{experiment-3-whisper}

The underlying GigaSpeech-derived candidate pool contained 227,898
timestamped speech segments in which \emph{up} occurred, of which
220,285 contained a V+\emph{up} instance, spanning 5,275 unique
V+\emph{up} types, before filtering to the final test set described in
the main text.

\begin{table}[H]

\centering{

\begin{tabular}{lrrrr}
\toprule
Component & Train pos & Train neg & Val pos & Val neg\\
\midrule
Encoder & 1,000 & 1,000 & 1,000 & 1,000\\
Decoder & 1,000 & 1,000 & 1,000 & 1,000\\
\bottomrule
\end{tabular}

}

\caption{\label{tbl-exp3-splits}Classifier training and validation split
sizes for Experiment 3 (Whisper). Positive examples are \emph{word\_up}
instances; negatives are random non-\emph{up} alphabetic tokens from the
same segment.}

\end{table}%

\begin{table}[H]

\centering{

\begin{tabular}{llrrlr}
\toprule
Model & Coef & Est. & Err. & 95\% CI & \% > 0\\
\midrule
Encoder & Intercept & 9.72 & 0.04 & {}[9.64, 9.81] & 100.0\\
Encoder & Freq & 0.06 & 0.04 & {}[-0.03, 0.14] & 91.0\\
Encoder & Predic. & -0.07 & 0.04 & {}[-0.16, 0.02] & 5.5\\
Encoder & Duration & -0.12 & 0.02 & {}[-0.16, -0.07] & 0.0\\
Encoder & Freq:Predic. & 0.01 & 0.04 & {}[-0.06, 0.09] & 64.3\\
Decoder & Intercept & 6.15 & 0.06 & {}[6.03, 6.28] & 100.0\\
Decoder & Freq & -0.18 & 0.06 & {}[-0.29, -0.06] & 0.1\\
Decoder & Predic. & -0.70 & 0.07 & {}[-0.83, -0.58] & 0.0\\
Decoder & Duration & 0.07 & 0.03 & {}[0.01, 0.12] & 99.5\\
Decoder & Freq:Predic. & -0.05 & 0.06 & {}[-0.16, 0.06] & 20.8\\
\bottomrule
\end{tabular}

}

\caption{\label{tbl-exp3-joint-final}Joint frequency × predictability
model at the final layer for Whisper encoder and decoder, controlling
for the duration of the \emph{up} audio segment (a proxy for phonetic
reduction). Estimates are from Bayesian linear mixed-effects models
(brms); \% \textgreater{} 0 indicates the percentage of posterior
samples with a positive estimate.}

\end{table}%

\begin{table}[H]

\centering{

\begin{tabular}{llrrl}
\toprule
Model & Predictor & EDF & F & p\\
\midrule
Encoder & Frequency & 13.43 & 81.33 & < 0.001\\
Decoder & Frequency & 21.17 & 9361.71 & < 0.001\\
Encoder & Predictability & 17.39 & 60.29 & < 0.001\\
Decoder & Predictability & 22.29 & 9258.61 & < 0.001\\
\bottomrule
\end{tabular}

}

\caption{\label{tbl-exp3-gam-smooth}GAM tensor-product smooth summary
for Whisper encoder and decoder, controlling for the duration of the
\emph{up} audio segment. EDF = estimated degrees of freedom; p = p-value
for the te(predictor, layer) smooth term.}

\end{table}%

\begin{figure}

\centering{

\pandocbounded{\includegraphics[keepaspectratio]{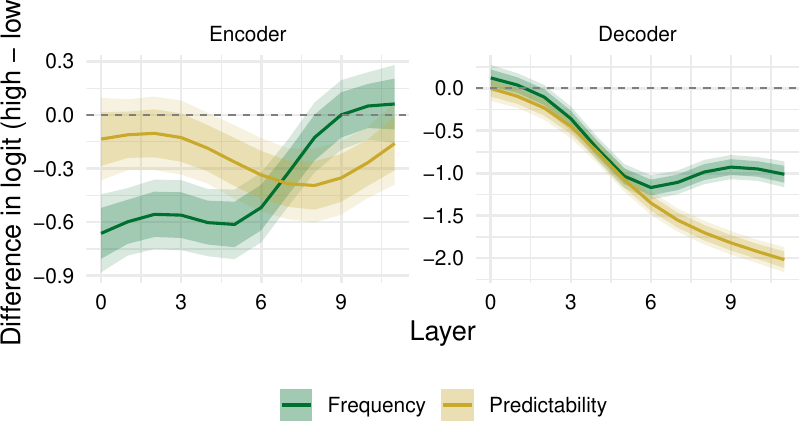}}

}

\caption{\label{fig-exp3-gam-diffs}Layer-by-layer difference in
GAM-predicted logit between high (90th percentile) and low (10th
percentile) predictor values, for Whisper encoder and decoder. Green =
frequency effect; yellow = predictability effect. Inner and outer shaded
ribbons show 80\% and 95\% confidence intervals derived from the GAM
standard error. A negative difference indicates that high-predictor
phrases yield a lower classifier logit (i.e., their representation of
\emph{up} is less similar to the preposition class).}

\end{figure}%

\clearpage

\end{document}